\documentclass[journal]{IEEEtran}

\usepackage{xcolor}
\usepackage{amsmath}
\usepackage{amssymb}
\usepackage{graphicx}
\usepackage{mathtools}
\usepackage{nicefrac} 
\usepackage[algo2e]{algorithm2e} 
\usepackage{algorithmic}
\usepackage{algorithm}
\usepackage{booktabs}

\usepackage{stackengine}
\usepackage{caption}
\usepackage{subfigure}

\DeclareMathAlphabet{\pazocal}{OMS}{zplm}{m}{n}

\newcommand{\Lb}{\pazocal{L}}
\usepackage{microtype}
\usepackage{graphicx}
\usepackage{mathtools}
\usepackage{amsmath,bm}
\usepackage{multirow}
\usepackage{subfigure}
\usepackage{booktabs} 
\usepackage{hyperref}
\usepackage{amsmath}
\usepackage[super]{nth}
\usepackage{stackengine}

\usepackage{amssymb}  
\usepackage{color}

\newcommand\mydots{\makebox[1em][c]{.\hfil.\hfil.}}

\begin{document}
\title{Low-rank Characteristic Tensor Density Estimation Part II: Compression and Latent Density Estimation}

\author{Magda~Amiridi,
        Nikos~Kargas,
        and~Nicholas D. Sidiropoulos,~\IEEEmembership{Fellow,~IEEE}
\thanks{
M. Amiridi and N.D. Sidiropoulos are with the Department of ECE, University of Virginia, Charlottesville, VA 22904. Author e-mails: (ma7bx,nikos)@virginia.edu.
N. Kargas was with the Department of ECE, University of Minnesota; he is now with Amazon, Cambridge, U.K. Author e-mail: karga005@umn.edu}}

\maketitle
\begin{abstract}
Learning generative probabilistic models is a core problem in machine learning, which presents significant challenges due to the \textit{curse of dimensionality}. This paper proposes a joint dimensionality reduction and non-parametric density estimation framework, using a novel estimator that can explicitly capture the underlying distribution of appropriate reduced-dimension representations of the input data. The idea is to jointly design a nonlinear dimensionality reducing auto-encoder to model the training data in terms of a parsimonious set of latent random variables, and learn a \textit{canonical} low-rank tensor model of the joint distribution of the latent variables in the Fourier domain. The proposed latent density model is non-parametric and ``universal'', as opposed to the predefined prior that is \textit{assumed} in variational auto-encoders. Joint optimization of the auto-encoder and the latent density estimator is pursued via a formulation which learns both by minimizing a combination of the negative log-likelihood in the latent domain and the auto-encoder reconstruction loss. We demonstrate that the proposed model achieves very promising results on toy, tabular, and image datasets on regression tasks, sampling, and anomaly detection.   
\end{abstract}


\IEEEpeerreviewmaketitle

\section{Introduction}
\noindent Accurate modeling of the multivariate structure of data based on observed data samples is one of the most fundamental topics in  machine learning. A model of the joint probability density function (PDF) of a data vector encodes the complete statistical properties of the data generative process and allows one to reason about data probabilistically, uncover the low-dimensional manifold the data is assumed to live on, and ultimately generate new data. PDF estimation serves as a building block in a wide variety of applications, such as image processing~\cite{kingma2018glow}, speech modeling~\cite{oord2016wavenet}, natural language processing~\cite{bowman2015generating}, and anomaly detection~\cite{zong2018deep}. Conventional density estimation methods, such as kernel density estimation (KDE)~\cite{silverman1986density} and Gaussian mixture models (GMMs)~\cite{mclachlan2004finite} are usually designed to fit target distributions directly in the data space  $\mathbb{R}^N$ and fall short in high-dimensions from both computational and statistical points of view due to the Curse of Dimensionality -- convergence slows down as the number of dimensions increases as a result of data sparsity in high-dimensional spaces. Real-world data often resides in a high-dimensional and complex feature space with only a limited amount of observed data being directly available.

Recently, the use of deep neural networks has led to substantial advances in this area. For example, generative adversarial networks (GANs)~\cite{goodfellow2014generative} can be trained to sample from very high-dimensional densities, but they do not support statistical inference or explicit density evaluation. On the other hand, variational auto-encoders (VAEs)~\cite{kingma2013auto} provide functionality for both (approximate) inference and sampling. VAEs assume a prior as a manually specified distribution (e.g., a simple isotropic Gaussian or mixture of Gaussians) and are trained  by minimizing  a  reconstruction  error and a divergence to force the variational posterior to fit the prior of the latent variables. However, the forced global structure in the latent space through the use of a manually specified prior may differ from the complex latent nature of the true data manifold. Thus, such simplistic assumptions may potentially harm the generalization of high dimensional data from low dimensional latent spaces. For example, it is observed that VAEs tend to generate blurry images, an effect that is usually attributed to the latent density mismatch problem ~\cite{dai2019diagnosing, rosca2018distribution}.
Finally, explicit neural models such as auto-regressive models~\cite{oord2016pixel} and flow-based models~\cite{dinh2014nice,dinh2016density} are designed to perform sampling and point-wise density evaluation. Despite their success, auto-regressive models generally suffer from slow sampling time~\cite{ho2019flow++} and inferior quality of samples compared to VAEs; but they are particularly useful for point-wise density evaluation. On the other hand, flow-based models, such as Real-NVP~\cite{dinh2016density} and Glow~\cite{kingma2018glow}, are efficient for sampling, but have inferior performance in evaluating the log-likelihood of the input compared to the auto-regressive ones. 

The goal of this paper is to introduce a class of probabilistic latent variable models for
unsupervised learning which is tailored for high dimensional datasets. The proposed class of models is non-parametric, and it learns the underlying distribution of latent representations of the input data in the Fourier domain. The proposed framework consists of two main components: an auto encoder network, through which a lower-dimensional latent representation of the input is sought; and a nonparametric density estimation module in the latent domain. The auto-encoder compresses redundancies in the data domain while preserving the essential information, and is used as a new feature representation space where we learn the data distribution. The auto encoder and the latent density are learned jointly via an optimization criterion that combines a data reconstruction loss and a negative log-likelihood regularization term over the latent representations of the training data.  

This is the second part of a two-part paper. The first part \cite{amiridi2020nonparametric} dealt with the density estimation problem in the native (``raw'') input data domain, showing that any joint density that is compactly supported and continuously differentiable can be well-approximated using a low-rank tensor model in the Fourier domain. A corollary of \cite{amiridi2020nonparametric} is that a finite separable mixture model (approximately) follows from  compactness of support and continuous differentiability. This interpretation enables an efficient and disciplined  sampling process. By introducing a low-rank tensor model in the Fourier domain via the Canonical Polyadic Decomposition (CPD)~\cite{Harshman1970}, a controllable approximation of the multivariate density is identifiable. The choice of tensor rank, number of Fourier coefficients, and the dimensionality of the latent space let us control the expressivity of the learned distribution. With respect to Part I \cite{amiridi2020nonparametric}, the key differences in this second part are the following:
\begin{itemize}
\item  Unlike Part I, which aimed to tackle the problem directly in the original $N$-dimensional space, probabilistic modeling in Part II is realized in a reduced-dimension latent space and the effects are translated back into the input space through the decoder mapping. Towards this end, a {\em joint} nonlinear dimensionality reduction and compressed density estimation framework is proposed in Part II. The joint approach boosts the flexibility, scalability, and statistical performance in terms of prediction (regression/detection) accuracy and sampling fidelity. 
\item Instead of the coupled tensor factorization approach adopted in Part I, Part II tackles joint density estimation as a {\em hidden tensor factorization} problem using maximum likelihood learning of the latent distribution's parameters.
\end{itemize}

\begin{figure}
\begin{center}
\includegraphics[width=0.9\linewidth]{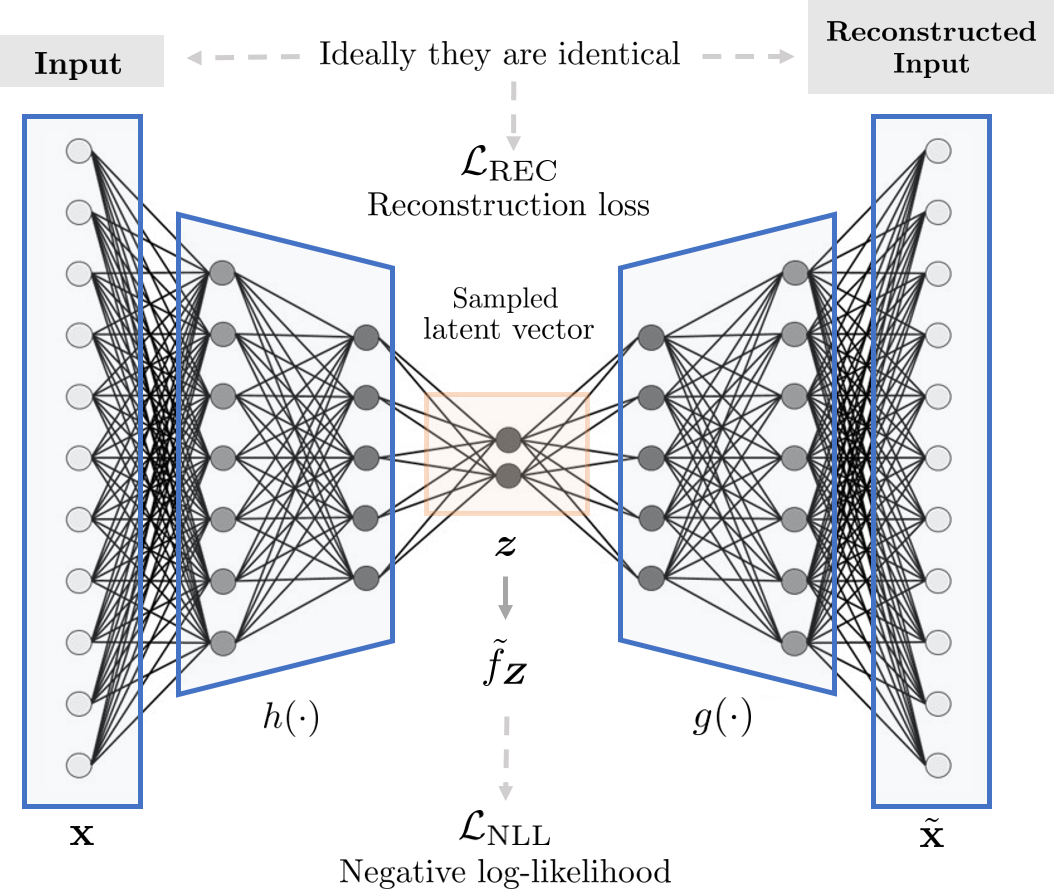}
\caption{Compressed Density Estimation: An  auto-encoder attempts to reproduce  the  input in the output layer by compressing it to fewer dimensions while retaining non-redundant information. The hidden layer becomes a bottleneck, forming a  lower-dimensional representation of the data, which is used to build a non-parametric density model. }
\label{CDE}
\end{center}
\end{figure}

A high-level overview of the proposed framework is shown in Figure~\ref{CDE}. A sneak peak of the expected performance of the proposed method is shown in Figure~\ref{vae-comp} where we use our model to learn the joint distribution of MNIST \cite{lecun1998mnist} images of $0$s and $8$s. With a suitable combination of hyper-parameters, the proposed density estimator offers considerable flexibility without sacrificing parsimony of representation. We showcase the promising results of the proposed model on benchmark image (MNIST, FMNIST \cite{xiao2017fashion}) and several tabular datasets on sampling, regression, and anomaly detection tasks, and on some toy but didactic examples for illustration.

\section{Background}
\subsection{Related work}
\label{sec:related_work}
Classic work on density estimation includes Gaussian Mixture Models, which are fragile to model mismatch due to their parametric nature, and introduce computational and estimation challenges in the high dimensional case. Conventional non-parametric models such as the Kernel Density Estimators become computationally intractable in high dimensions, since the number of parameters grows exponentially with the number of dimensions~\cite{scott1991feasibility}.

Recently, the use of deep neural networks has led to significant advances in modeling modern complex and high-dimensional data. Auto-encoders (AE) enjoy  a  remarkable  ability  to  learn data  representations. Auto-encoder networks  such as VAEs~\cite{kingma2013auto} and GANs~\cite{goodfellow2014generative} learn latent representations of very high-dimensional data such as images or videos. However, GANs only support sampling, but not inference or density estimation. VAEs assume that high-dimensional data can be modeled as lying on or near a low-dimensional, nonlinear manifold which they approximate by learning nonlinear mappings while encouraging a global structure in the latent space through the use of a specified prior distribution. However, specifying the prior distribution may prevent them from faithfully representing the true data  manifold. It was shown in \cite{hoffman2016elbo} that choosing a too simplistic prior could lead to over-regularization and, as a consequence, very poor hidden representations. A key advantage of our approach is that we introduce a non-parametric density model into the latent space, which by virtue of uniqueness of low rank tensor decomposition comes with approximation guarantees. This approach can yield a more accurate model of the data manifold, as we will see. A conceptually similar approach was proposed in \cite{zong2018deep} and applied for unsupervised anomaly detection, the key difference being that the density of low-dimensional representations was modelled using a GMM, which is far more restrictive and does not come with identification guarantees.

\begin{figure}[!t]
\begin{center}
\subfigure[\textbf{From left to right} : $(F=4, K=1), (F=4, K=3), (F=4, K=5)$]{ \includegraphics[width=0.9\linewidth]{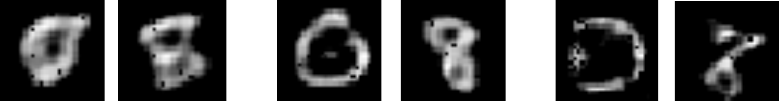}} \quad
\subfigure[\textbf{From left to right} : $(F=2, K=3), (F=4, K=3), (F=8, K=3)$]{ \includegraphics[width=0.9\linewidth]{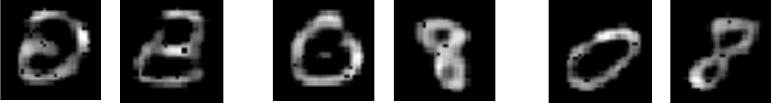}} \quad \quad
\end{center}
\caption{Sneak peek: Demonstration of generated MNIST samples trained on images of $0$s and $8$s using the proposed CDE model. We train the model on different values of $F$ and $K$ to show that only a few parameters are needed to come close to the ground-truth. Increasing $K$ generates sharper digits, while increasing $F$ better differentiates the samples of the  digits.}
\label{vae-comp} 
\end{figure}

Other classes of generative models include the Real-valued Neural Autoregressive Distribution Estimator (RNADE)~\cite{uria2013rnade} and its discrete version MADE~\cite{germain2015made}, which is among the best performing neural density evaluation methods and has shown great potential in scaling to high-dimensional distribution evaluation problems. These so-called autoregressive models decompose the joint density as a product of one-dimensional conditionals of increasing conditioning order, and model each conditional density with a parametric model. Normalizing Flows (NF)~\cite{rezende15} models, on the other hand,  start with a base density e.g., standard Gaussian, and stack a series of invertible transformations with tractable Jacobian to approximate the target density. Masked Autoregressive Flow (MAF)~\cite{papamakarios2017masked} is a type of NF model where the transformation layer is built as an autoregressive neural network. Finally, Gaussianization flows (GF)~\cite{meng2020gaussianization} build upon rotation-based iterative Gaussianization. These methods do not construct an explicit joint PDF model, but rather serve for point-wise density evaluation. That is, for any given input vector (realization), they output an estimate of the density evaluated at that particular input vector. 

\subsection{Notation}
\label{sec:Notation}
In this paper, we use $\mathbf{x}$, $\mathbf{X}$, $\underline{\mathbf{X}}$ for vectors, matrices and tensors respectively. We use the notation $\mathbf{x}(k)$, $\mathbf{X}(:,k)$, ${\underline{\mathbf{X}}(:,:,k)}$ to refer to a particular element of a vector, a column of a matrix and a slab of a tensor. Symbols $\|\mathbf{x}\|_2$ and  $\|\mathbf{X}\|_F$ correspond to $L_2$ norm and Frobenius norm. Symbols $\circ$, $\circledast$, $\odot$ denote the outer, Hadamard and Khatri-Rao product respectively. The set of integers $\{1,\ldots, N\}$ is denoted as $[N]$. We use the notation FC (a, b, c) to describe a fully-connected layer with a input neurons and b output neurons activated by function c.


\subsection{Canonical Polyadic Decomposition}
In this section, we briefly introduce basic concepts related to tensor decomposition. A $D$-way tensor ${\underline{\boldsymbol{\Phi}} \in \mathbb{C}^{K_1 \times K_2 \times \cdots \times K_D}}$ is a multidimensional array whose elements are indexed by $D$ indices. Any tensor can be decomposed as a sum of $F$ rank-$1$ tensors, i.e.,
\begin{equation}
\underline{\bm{\Phi}}  = [\![\boldsymbol{\lambda}, \mathbf{A}_1,\mathbf{A}_2, \ldots,\mathbf{A}_D]\!] = \sum_{f=1}^F\boldsymbol{\lambda}(f) \mathbf{a}^{1}_f \circ \mathbf{a}^{2}_f \circ \cdots  \circ \mathbf{a}^{D}_f,
\label{eq:cpd}
\end{equation}
where $\mathbf{A}_d = [ \mathbf{a}_1^d, \ldots, \mathbf{a}_F^d] \in \mathbb{C}^{K_D\times F}$ and constraining the columns ${\mathbf{A}_n(:,f)}$ to have unit norm, the real scalar ${\boldsymbol{\lambda}}(f)$ absorbs the $f$-th rank-one tensor’s scaling.
\medskip
A particular element of the tensor is given by 
\begin{equation}
\underline{\bm{\Phi}}(k_1,k_2,\ldots,k_D) = \sum_{f=1}^F\boldsymbol{\lambda}(f)\prod_{d=1}^D \mathbf{A}_d(k_d,f).
\end{equation}
When $F$ is minimal, it is called the rank of $\underline{\bm{\Phi}}$, and the decomposition is called Canonical Polyadic Decomposition (CPD) \cite{hitchcock1927expression,Harshman1970}. CPD is a powerful model that can parsimoniously represent the high-order interactions among multi-way data exactly or approximately leading to significant reduction in the number of parameters. A key property of the CPD  is that the rank-$1$ components are unique under mild conditions. For learning latent variable statistical models the uniqueness of tensor decomposition often translates to identifiability, that is the existence of a unique set of parameters that can be consistent with what we have observed. See~\cite{sidiropoulos2017tensor} for a tutorial overview and detailed identifiability results.

\section{Compressed-domain Density Estimation}

We consider the problem of general-purpose modeling of a high-dimensional continuous joint distribution $f_{{\bm X}}$ of an $N$-dimensional random vector ${\bm X}$ when $N$ is large.
Given a dataset $\mathcal{D}$ of $M$ i.i.d.  realizations in  the $N$-dimensional  observable  space $\mathcal{D} = \left\{ {\mathbf x}_m  \right\}_{m=1}^M$, we typically wish to perform maximum likelihood learning of its parameters, i.e., to minimize the Negative Log-Likelihood (NLL)
\begin{equation}
\mathcal{L}_{\rm NLL}= - \frac{1}{M} \sum\limits_{m=1}^{M}\text{log}\Big({{{f}}_{{\bm X}}({{\mathbf{x}}_{m}})}\Big).
\end{equation}
In general, $\mathcal{L}_{\rm NLL}$ is difficult to compute or differentiate directly, since the density ${f}_{{\bm X}}$ can be analytically and computationally intractable. One can address this issue and evade the curse of dimensionality by using a mapping $h$ to encode the input data samples ${\mathbf x}_m \in \mathbb{R}^N$ into much lower dimensional representations ${\mathbf z}_m \in \mathbb{R}^D$, with $D\ll N$, in the latent space. 

In this work, we propose a joint dimensionality reduction (DR) and density estimation framework where the DR part is carried out through learning an auto-encoder: 
\begin{equation*}\label{eq:AE}
\text{Auto-encoder:}~~ \mathbf{x} \overset{{h}}{\mapsto} \mathbf{z} \overset{{g}}{\mapsto} \tilde{\mathbf{x}}.
\end{equation*} 
Here, ${h}$ and ${g}$ denote the encoder and the decoder, respectively, and $\tilde{\mathbf{x}}$ is the reconstruction of $\mathbf{x}$. The mapping ${h}:\mathbb{R}^N \rightarrow \mathbb{R}^D$ can be viewed as nonlinear dimensionality reduction, and the low-dimensional ${\mathbf{z}}={h}({\mathbf{x}})$ as the bottleneck representation of the observed vector ${\mathbf{x}}$. We approximate the latent domain distribution $f_{\bm Z}$ using the  non-parametric density estimation framework in Part I of this work ~\cite{amiridi2020nonparametric}. The density estimation framework relies on the decomposition of a $D$-way tensor of leading Fourier series coefficients through CPD. 
Part I has shown that this model is quite general, in that it can approximate any multivariate compactly supported density as long as its Fourier coefficients decay sufficiently fast, and under certain conditions it can identify the true latent model. 

The choice of tensor rank, number of Fourier coefficients, and the dimensionality of the bottleneck representation are used to control the expressivity of the model. Here we propose to {\em jointly learn} the auto-encoder and the parameters of the density model.  The combination of an auto-encoder and density estimation takes advantage of their synergistic strengths. Auto-encoders can compress input data to fewer dimensions while retaining non-redundant information, while density estimation works best in lower-dimensional spaces. The proposed framework can be used for missing data imputation and as a generative model.
 
\medskip
\noindent \textbf{Missing data imputation}: Assume that for a given data sample $\mathbf{x}$, we observe a subset of its values denoted as ${\mathbf x}_O$ and ${\mathbf x}_M$ is the part that we do not observe. Data imputation can be performed by clamping the observed dimensions ${\mathbf x}_O$ to their values and maximizing log-likelihood with respect to the missing dimensions ${\mathbf x}_M$
\begin{align} 
& \max_{{\mathbf{x}}_M}  \;  
\log \left({f}_{{\bm Z}}\big({h}({\mathbf{x}}_O,{\mathbf x}_M)\big)\right).
\label{di}
\end{align} 

\textbf{Data sampling}: With ${g}$ given, we can draw a realization of the random vector ${\bm{Z}}$ in the $D$-dimensional latent space from $f_{\bm Z}$, and back transform to a sample in the original $N$-dimensional space by its inverse image as 
\begin{equation}
{\mathbf z} \sim{f}_{{\bm Z}},~ \tilde{\mathbf x} = g ({\mathbf z}).
\end{equation}
Similar approaches such as VAEs pose a stochastic condition on the latent variables to comply with a fixed prior distribution ${f}_{{\bm Z}}$ over a low-dimensional latent space: 
\begin{equation*}
\text{VAE:}~~\mathbf{x} \overset{{h}}{\mapsto} \mathbf{z} \overset{{g}}{\mapsto} \tilde{\mathbf{x}}, ~ \mathbf{z}  \sim {f}_{{\bm Z}}({\mathbf{z}}).
\end{equation*}
The generative process of the VAE is carried out as
\begin{equation*}
{\mathbf z} \sim{f}_{{\bm Z}},  {\mathbf x}\sim p_\theta({\mathbf X}|{\mathbf Z}={\mathbf z})    
\end{equation*}
where a stochastic decoder 
\begin{equation*}
D
_\theta(\mathbf z) ={\mathbf x}\sim p_\theta({\mathbf x}|{\mathbf z}) = p({\mathbf X}|g ({\mathbf z}))    
\end{equation*}
links the latent space to the input space through the likelihood distribution $p_\theta$. This may be limiting in case this predefined prior does not match the structure of the true data manifold, leading to a less accurate model. Our approach is fundamentally different as we avoid prior distribution matching between the variational posterior and the prior, but instead propose jointly learning a non-parametric density estimator in the latent space. Most importantly, our approach produces better samples than VAEs, as we will see. In the following sections, we give a detailed description of the two main components of our framework and the optimization procedure.

\subsection{Compression Network}
The first component of our framework seeks a non-linear mapping ${h}$ to project high-dimensional input samples into a low-dimensional space. In the dimensionality reduction process, discarding some dimensions inevitably leads to information loss. We wish to preserve the available information as much as possible, and to this end we minimize the empirical approximation of the mean squared error
\begin{equation}
{\rm{MSE}}:=\int_{\mathbb{R}^N}{\|{\bm x}-{g}({h}({\bm x}))\|}_2^2 {f}_{\bm X}({\bm x})d{\bm x}.
\end{equation}
Auto-encoders learn a function by fine-tuning the parameters of a feed-forward Deep Neural Network (DNN) in such a way that the reconstruction error is minimized when back projected with another feed-forward DNN. These networks need to be specified a-priori, in terms of the number of layers and neurons. In this work, we use the rectified linear unit (ReLU) activation function \cite{nair2010rectified} while the rest of the parameters such as the width of each layer and the depth of the network are adjusted according to $M, N$.

Although we proceed with simpler networks, other types of networks (e.g., convolutional neural networks \cite{lecun1998gradient, krizhevsky2017imagenet}) can also be used. Let ${h}(\cdot;{\boldsymbol{\theta}_h})$ and $g(\cdot;{\boldsymbol{\theta}_g})$ be DNNs and $\boldsymbol{\theta}_h$, $\boldsymbol{\theta}_g$ collect the encoder and decoder network parameters, i.e., the weights and bias terms at each hidden layer. Given a finite set of samples $M$, the empirical reconstruction loss can be computed as
\begin{equation}
\Lb_{\text{REC}}:=\frac{1}{M}\sum_{m=1}^M{||{\mathbf x}_m-{g}({ h}({\mathbf x}_m;{\boldsymbol{\theta}_h});{\boldsymbol{\theta}_g})||}_2^2.
\end{equation}
The reconstruction loss is typically minimized using Stochastic Gradient Descent (SGD).
\subsection{Latent Density Estimation Network} 
\label{sec:Main_method}
The second component of our framework is a non-parametric density estimation model~\cite{amiridi2020nonparametric}. 
The key difference is that  we propose joint dimensionality reduction and density modeling in the reduced-dimension latent space so that we capture the bottleneck layer distribution, whereas~\cite{amiridi2020nonparametric} aimed to tackle the problem directly in the original $N$-dimensional space.  This combination is crucial for enhanced performance and scalability. Additionally, instead of coupled tensor factorization, we consider an alternative algorithmic approach by formulating density estimation as a hidden tensor factorization problem. 



Let us consider the multivariate joint PDF ${{f}}_{{\bm Z}}$ of a $D$-dimensional random vector ${\bm Z}$ with its support contained within the hypercube $S =[0,1]^D$. Then, the joint PDF can be represented by a multivariate Fourier series
\begin{equation}
\begin{aligned}
f_{{\bm Z}}({\mathbf{z}})= {\sum_{k_1=-\infty}^\infty} \cdots{\sum_{k_D=-\infty}^\infty}{\Phi}_{\bm Z}[{\boldsymbol{k}}]e^{-j2\pi {\mathbf k}^T{\mathbf z}},\\ \end{aligned}
\end{equation}
where 
\begin{equation*}
{\Phi_{\bm Z}}[{\mathbf{k}}] = \Phi_{\bm Z}(\boldsymbol{\nu})\big\rvert_{{\boldsymbol{\nu}}=2\pi \mathbf{k}}, \mathbf{k} = [k_1,\ldots,k_D]^T
\end{equation*}
 and $\Phi_{Z}$ is the Characteristic Function (CF). The multivariate characteristic function $\Phi_{Z}:{\mathbb{R}}^D\rightarrow{\mathbb{C}}$ is defined as 
 \begin{equation*}
 \Phi_{{\bm Z}}({\bm \nu}) = E\left[e^{j {\bm \nu}^T {\bm Z}}\right].    
 \end{equation*}
  Similar to the PDF $f_{{\bm Z}}$, its corresponding CF ${\Phi}_{{\bm Z}}$ contains complete information about the distribution of ${\bm Z}$, i.e., the PDF and the CF have a bijective relationship – one being the Fourier transform of the other. When the underlying PDF is sufficiently differentiable in all variables, $f_{{\bm Z}}$ can be approximated by a truncated multivariate Fourier series with cutoffs $K_1,\ldots,K_D$ i.e., 
\begin{align}
\label{pdf_exp}
\tilde{f}_{{\bm Z}}({\bm{z}})={\sum_{k_1=-{K_1}}^{K_1}}\ldots{\sum_{k_D=-K_D}^{K_D}}{{{\Phi}}}_{\bm Z}[{\bm{k}}]e^{-j2\pi {\bm k}^T{\bm z}}.
\end{align} 
The smoother the underlying PDF the faster the convergence rate and the smaller the approximation error. 

\begin{figure}
\centering
\includegraphics[width=0.25\textwidth]{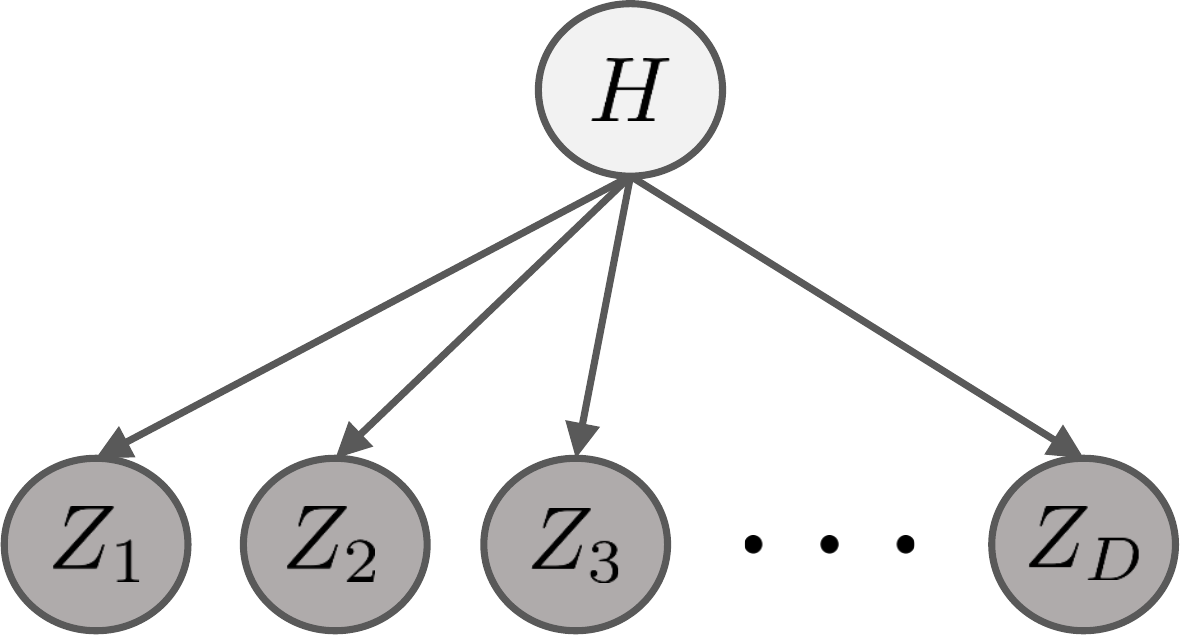}
\caption{Our approach yields a generative model of the latent density, from which it is very easy to sample from. This is because $f_{\bm Z}$ can be interpreted as a mixture of $F$ product distributions i.e., admits a latent variable naive Bayes interpretation.}
\label{fig:nb2}
\end{figure}

For any $p \in \mathbb{N}$, If the partial derivatives $\frac{\partial^{\theta_1}}{\partial z_1^{\theta_1}} \cdots \frac{\partial^{\theta_D}}{\partial z_D^{\theta_D}} f_{{\bm Z}}({\mathbf z})$ of $f(\cdot)$ exist and are absolutely integrable for all $\theta_1,\ldots,\theta_D$ with $\sum_{n=1}^D \theta_n \leq p$ then the rate of decay of the magnitude of the ${\mathbf k}$-th Fourier coefficient $|{{\Phi}_{{\bm Z}}}[{\mathbf k}]|$ obeys \cite{plonka2018numerical}
\begin{equation*}
|{\Phi_{{\bm Z}}}[{\mathbf{k}}]|=~\mathcal{O}{\bigg(}{{\frac{1}{1+\|\mathbf{k} \|_2^p}}}{\bigg)}.
\end{equation*}  The worst-case approximation error is bounded by
\begin{equation*}
\|f_{\bm Z}- \tilde{f}_{{\bm Z}} \|_{\infty}\leq C  \sum_{d=1}^D \frac{\omega_d \left(\frac{\partial^{\theta_d}}{ {\partial z_{d}^{\theta_d}}} f_{\bm Z},\frac{1}{1+K_d} \right)}{{(1+K_d)}^{\theta_d}}, 
\end{equation*}
where $C = C_2 \left(1+ C_1  \prod_{d=1}^D \log K_d\right)$, $C_1,C_2$ are constants independent of $f_{\bm Z}$ and the $K_d$'s and 
\begin{equation}
\begin{aligned}
\omega_j(f_{\bm Z},\delta):= \mathrel{\stackunder{$\displaystyle\text{sup}$}{
\stackunder{$\scriptstyle {\left|z_j-z_j'\right|\leq \delta}$}}}
|f_{\bm Z}(z_1,\mydots,z_j,\mydots,z_D) \\-f_{\bm Z}(z_1,\mydots,z_j',\mydots,z_D)|
\end{aligned}
\end{equation}
measures the smoothness of $f_{\bm Z}$ for each component ${j\in [D]}$~\cite{mason1980near},~\cite[Chapter~23]{handscomb2014methods}.
Note that we can represent the truncated Fourier coefficients using a $D$-way tensor $\underline{\boldsymbol{\Phi}}$ where
\begin{equation}
\underline{\boldsymbol{\Phi}}(k_1, \ldots, k_D) = \Phi_{\boldsymbol{Z}}[\mathbf{k}].
\end{equation} 
For simplicity we will assume that $K_1=\cdots=K_D=K$. Orthogonal series PDF approximation using a truncated sum of basis functions (e.g., trigonometric, polynomial, wavelet) becomes computationally intractable in high dimensions, since the number of parameters (tensor elements) grows exponentially with the number of dimensions. To reduce the number of parameters, we introduce a low-rank parameterization of the coefficient tensor obtained by truncating the multidimensional Fourier series~\cite{amiridi2020nonparametric} which reduces the number of parameters from $O(K^D)$ to $O(DKF)$. Introducing the rank-$F$ CPD we have
\begin{equation}
\underline{{{\bm{\Phi}}}}({k_1,\ldots,k_D})=\sum_{f=1}^F p_H(f) \prod_{d=1}^D\Phi_{Z_d|H=f}(k_d|f).
\end{equation}
By linearity and separability of the multidimensional Fourier transformation, applied to rank-one components, $ {\tilde{f}}_{{\bm Z}}(\mathbf{z})$ can be written in the following form 
\begin{align*}
\tilde{f}_{{\bm Z}}& (\mathbf{z}) ={\sum_{k_1=-{K_1}}^{K_1}}\ldots{\sum_{k_D=-K_D}^{K_D}}{\Phi}_{\bm Z}[{\mathbf{k}}]e^{-j2\pi \mathbf{k}^T  \mathbf{z}} \\
&=\sum_{f=1}^F {\underbrace{p_H(f)}_{{\boldsymbol{\lambda}(f)}}}\prod_{d=1}^D \sum\limits_{k_d=-K}^K{\underbrace{\Phi_{Z_d|H=f}(k_d|f)}_{\mathbf{a}_d^f(K+1+k_d)}}{\underbrace{e^{-j2\pi k_d z_d}}_{\mathbf{b}_d(K+1+k_d)}}\nonumber \\
&=\sum_{f=1}^F {\boldsymbol{\lambda}(f)}\prod_{d=1}^D \mathbf{A}_d(:,f)^T \mathbf{b}_d.\nonumber
\end{align*}
The above joint PDF $f_{{\bm Z}}$ model can be interpreted as a mixture of $F$ product distributions, i.e., there exists a `hidden' random variable $H$ taking values in $\left\{1,\ldots,F\right\}$ that selects the operational component of the mixture, and given $H$ the random variables $Z_1,\ldots,Z_D$ become independent (Fig.~\ref{fig:nb2}). Then, given $\mathbf{z}$, we can compute the likelihood using
\begin{align*}
{\widehat{f}}_{{\bm Z}}({\mathbf z}) &= ( \mathbf{b}_1^T \mathbf{A}_1 \circledast \cdots \circledast \mathbf{b}_D^T \mathbf{A}_D )^T \boldsymbol{\lambda}\\ &= (\circledast_{d=1}^D \mathbf{b}_d^T \mathbf{A}_d) {\boldsymbol{\lambda}}.
\end{align*}
The complexity of computing the likelihood of a data point $\mathbf{z}$ is $O(DKF)$.
Tensor methods are commonly used to establish that the parameters  of  a generative model can be identified  given higher order moments. This generative model theoretically has enough flexibility to capture highly complex distributions such as image manifolds. According to this model, a sample of the multivariate latent distribution can be generated by first drawing $H$ according to $p_H$ and then independently drawing samples for each variable $Z_d$ from the conditional PDF $f_{Z_d|H}$. 
\subsubsection{Maximum Likelihood Estimation}
The above analysis suggests fitting a low-rank CPD model on the $D$-way Fourier series coefﬁcient tensor $\underline{\bm{\Phi}}$. To this end, we build a generative probabilistic model that assigns high probability to the transformed observed samples. 
We propose fitting the Fourier tensor coefficients indirectly on the latent space representation of the training data. Let us define matrices  $\mathbf{B}_d \in \mathbb{C}^{K \times M}$ as 
\begin{equation*}
{\mathbf{B}_d(K+1+k_d,m) = e^{-j2\pi k_d \mathbf{z}_m(d)}}.
\end{equation*}
Given $M$ samples, we define the following NLL cost term 
\begin{equation} 
\begin{aligned}
\mathcal{L}_{\rm NLL}&:=-{\frac{1}{M}}\sum\limits_{m=1}^{M}\text{log}\Big({{\hat{f}}_{{\bm Z}}({{\bm z}_{m}})}\Big) \\
&=-{\frac{1}{M}}\sum\limits_{m=1}^{M}\text{log}\Big((\circledast_{d=1}^D (\mathbf{B}_d(:, m)^T \mathbf{A}_d)) {\boldsymbol{\lambda}}\Big),
\end{aligned}
\end{equation}
where $\bm{A}_d(K+1+k_d,h)$ holds $\Phi_{Z_d|H=h}[k_d]$, $\boldsymbol{\lambda}(f)$ holds $p_H(f)$. Note that we do not  instantiate the full Fourier coefficient tensor
but rather recover it implicitly, by minimizing the NLL term.

We can further restrict the model and reduce its learnable parameters by $50\%$ by noticing that each column of the factor matrix $\mathbf{A}_d$ holds a valid characteristic function which is by definition conjugate symmetric around the origin, and equal to one at the origin, i.e., 
\begin{align*}
\mathbf{A}_d(K+1,:)&=\mathbf{1}^{T},\text{ and }\\
\mathbf{A}_d(K+1+k,:)&= \mathbf{A}^*_d(K+1-k,:),\\
\quad \quad \quad \quad \quad \quad & k \in[K], d \in [D]. 
\end{align*}


\begin{algorithm}
\caption{CDE (Projected - SGD)}
\label{alg:FSA_HTF_SGD}
\begin{algorithmic}
\STATE {{\bfseries Input:} $\mathbf{Z}, \mathbf{Z}_{\rm val}, F, K, D, {M}_{\rm batch}$}
\STATE {\textbf{Initialize }$\boldsymbol{\lambda},\{\mathbf{A}_n \}_{n=1}^D$},$\boldsymbol{\theta}_{g}$, $\boldsymbol{\theta}_{h}$ 
\REPEAT 
\STATE { Sample ${M}_{\rm batch}$ data points}
\STATE Update network parameters via SGD\\
\FOR{$d=1$ {\bfseries to} $D$}
\STATE Update $\mathbf{A}_d$ via SGD\\
\ENDFOR
\STATE Update $\boldsymbol{\lambda}$
\STATE Project $\boldsymbol{\lambda}$  onto the probability simplex
\STATE Compute ${\Lb_{\rm{NLL}}+\Lb_{\text{rec}}}$ using  $\mathbf{Z}_{\rm val}$
\UNTIL{{$\rm{max_{iter}}$ is reached or ${\Lb_{\text{NLL}}+\Lb_{\text{rec}}}$ stops diminishing} }
\end{algorithmic}
\label{alg:CDE_HTF_SGD}
\end{algorithm}
\subsection{Optimization Procedure}

By  the  above reasoning, instead of using decoupled two-stage training we suggest the following overall joint DR and density estimation optimization problem which we tackle by stochastic gradient descent 
\begin{equation*}
\begin{aligned}
&\min_{\boldsymbol{\theta}_h,\boldsymbol{\theta}_g, \{ \mathbf{A}_d\}_{d=1}^D, \boldsymbol{\lambda}} \frac{1}{M}\sum_{m=1}^{M} \Big( \|{\mathbf x}_m- g (h({\mathbf x}_m;\boldsymbol{\theta}_h); \boldsymbol{\theta}_g)\|^2 -  \nonumber\\
& - \mu \log \Big((\circledast_{d=1}^D (\mathbf{B}_d(:, m)^T \mathbf{A}_d)) {\boldsymbol{\lambda}}\Big) \Big) + 
\sum_{d=1}^D \rho \| \mathbf{A}_d \|_F^2 \nonumber\\	
&\text{s.t. } \boldsymbol{\lambda}\geq \mathbf{0}, {\mathbf{1}^{T}\boldsymbol{\lambda}= 1}, \nonumber \\
& \quad \mathbf{A}_d(K+1,:)=\mathbf{1}^{T}, \\
&  \quad \mathbf{A}_d(K+1+k,:) = \mathbf{A}^*_d(K+1-k,:).
\label{eq: formulation}
\end{aligned}
\end{equation*}
The optimization criterion that guides CDE consists of three terms: the reconstruction loss of the AE, NLL of the density estimation component, and Frobenius norm regularization. In the above formulation, $\mu\geq 0$ is a regularization parameter which balances the reconstruction error versus the maximum likelihood estimation. The number of coefﬁcients $K$ controls the desired smoothness of the joint density, while the number of latent dimensions $D$ and the rank $F$ control the expressivity.  

We refer to this approach as Compressed-domain Density Estimation with Hidden Tensor Factorization (CDE-HTF). Figure~(\ref{CDE}) presents the network structure corresponding to the final joint problem formulation. 
We solve the proposed optimization problem using projected Stochastic Gradient Descent (SGD). We initialize the Fourier tensor-related parameters using random initialization, while for $\boldsymbol{\theta}_g$ and $\boldsymbol{\theta}_h$, it was empirically observed that auto-encoder pre-training was most effective. At each step we update $\boldsymbol{\theta}_g$, $\boldsymbol{\theta}_h$, factors $\mathbf{A}_d$ and $\boldsymbol{\lambda}$ simultaneously by first sampling a batch of size ${M}_{\rm batch}$ and taking a gradient step. After that, we project $\boldsymbol{\lambda}$ to the probability simplex. For the termination of the algorithm we compute the cost function on a validation set and stop if a number of maximum iterations has been reached or the log-likelihood has not improved in the last $T$ iterations. The full procedure is shown in Algorithm~\ref{alg:CDE_HTF_SGD}. 
\section{Experimental Results}
In this section, we evaluate the proposed approach using various datasets and evaluation criteria, ranging from sampling of toy 3-D examples to real MNIST and Fashion-MNIST images, and regression and anomaly detection tasks using standard  tabular datasets from the UCI database. We compare with density estimation and anomaly detection baselines from the deep learning literature, including standard VAEs, Real-NVP, MAF, MADE and GF for reference. 
\subsection{Toy Datasets}
\begin{figure*}
\begin{center}
\subfigure[Latent space $\mathcal{Z}$ for the Swiss-roll dataset.]{\includegraphics[width=0.28\linewidth]{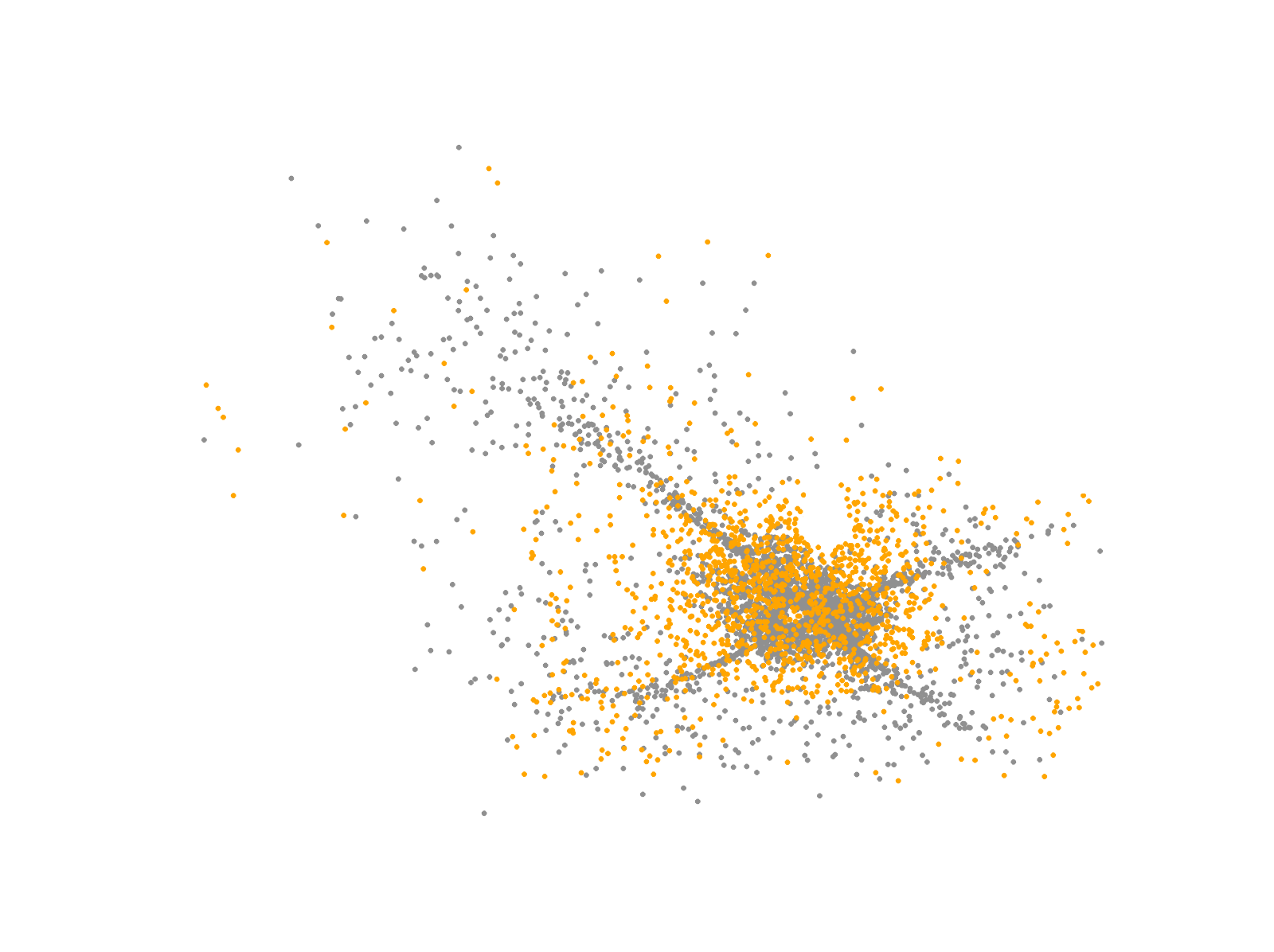}} \quad
\subfigure[Latent space $\mathcal{Z}$ for the S-Curve dataset.]{\includegraphics[width=0.28\linewidth]{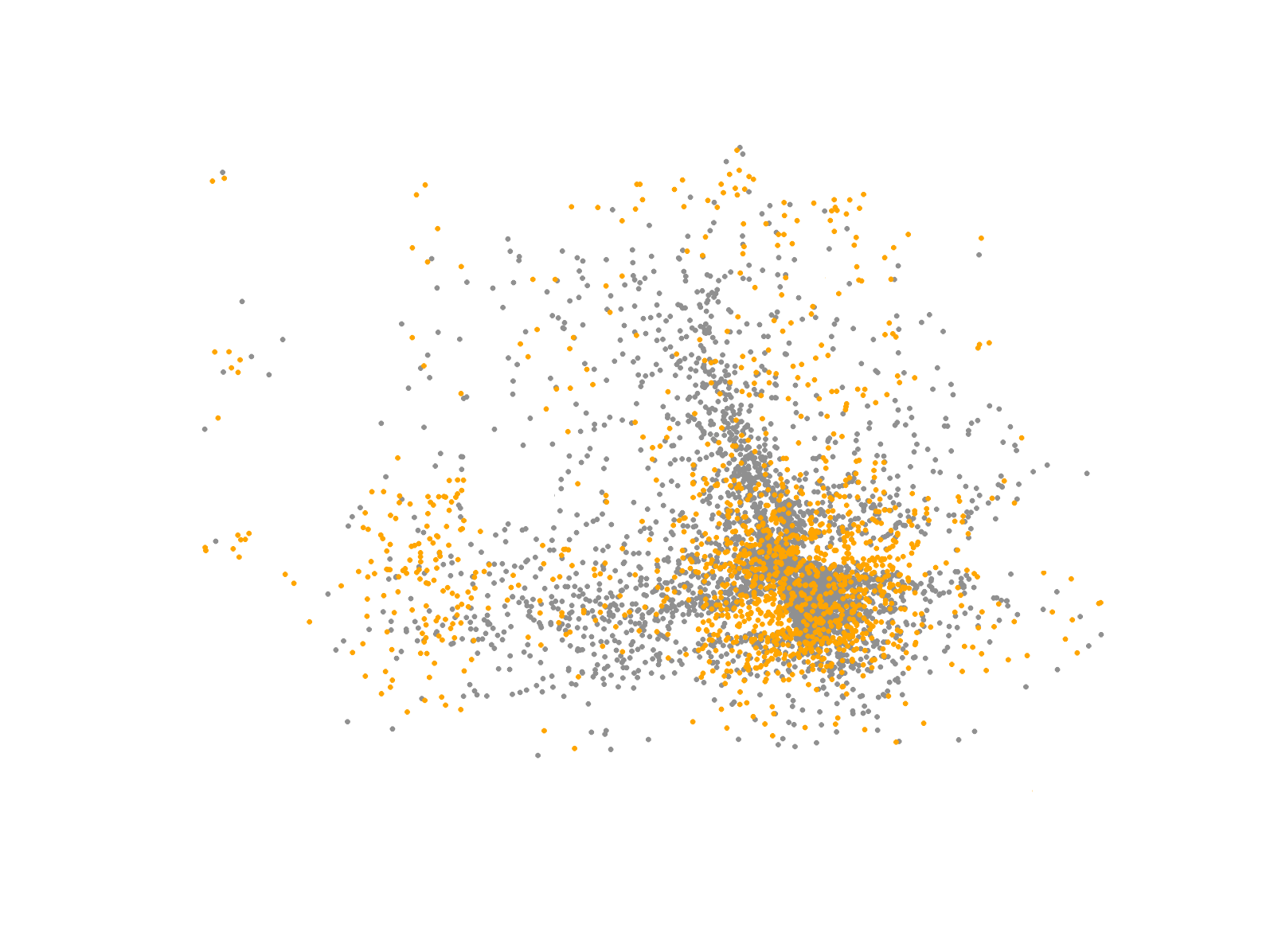}} \quad
\subfigure[Latent space $\mathcal{Z}$ for the Fish bowl dataset.]{\includegraphics[width=0.28\linewidth]{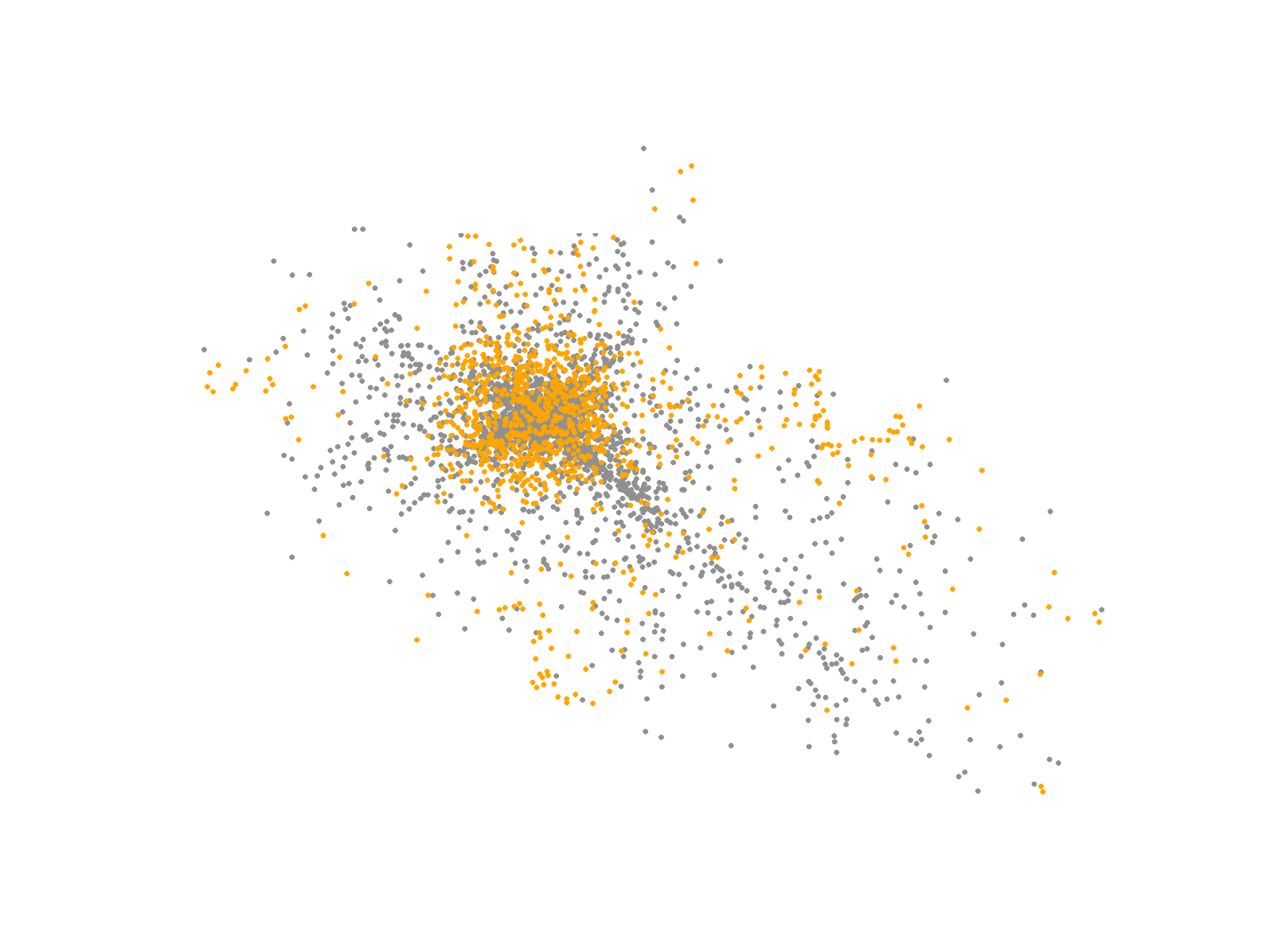}}

\subfigure[Data space $\mathcal{X}$for the Swiss-roll dataset. ]{\includegraphics[width=0.28\linewidth]{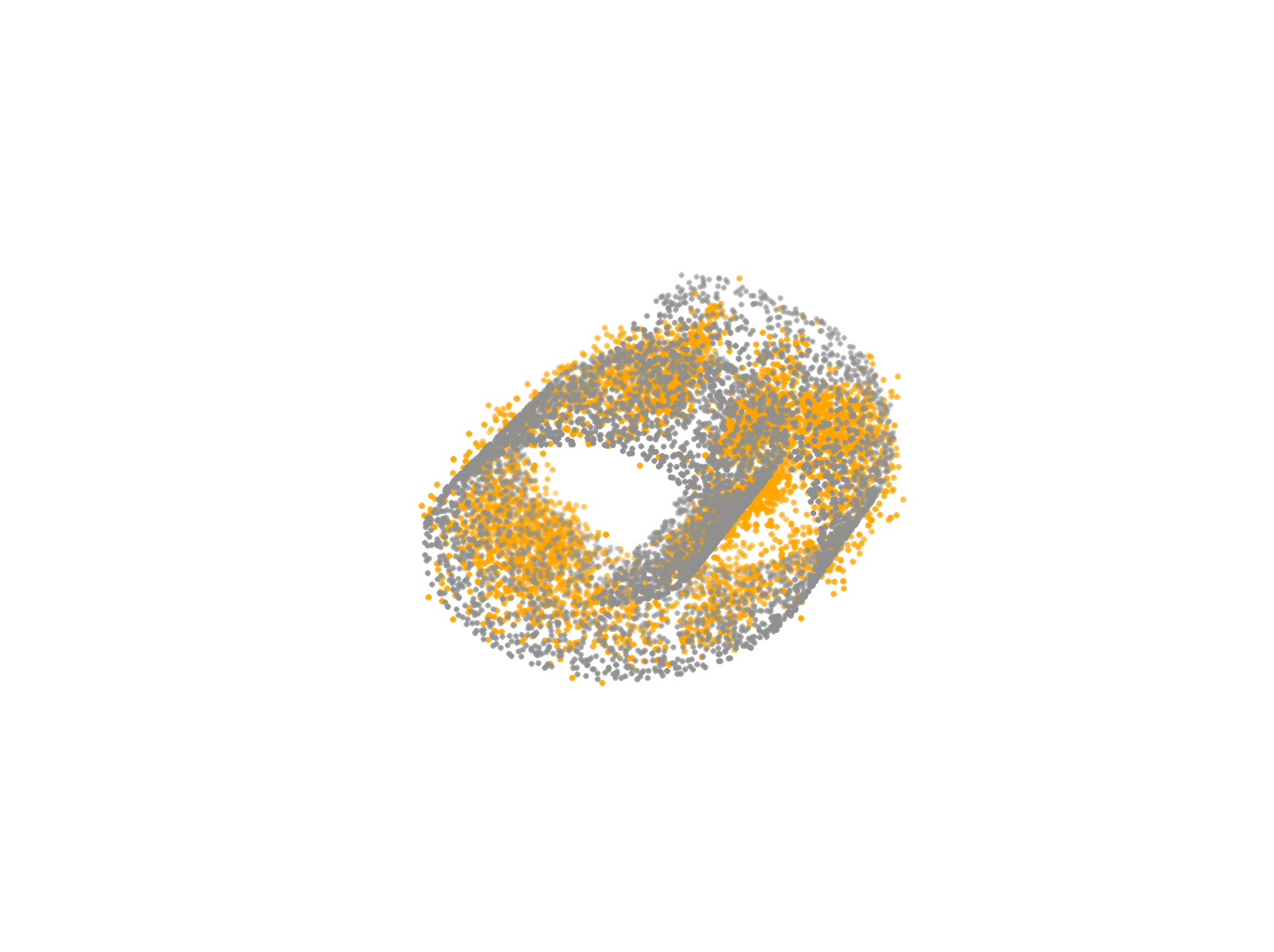}}
\quad
\subfigure[Data space $\mathcal{X}$ for the S-Curve dataset.]{\includegraphics[width=0.28\linewidth]{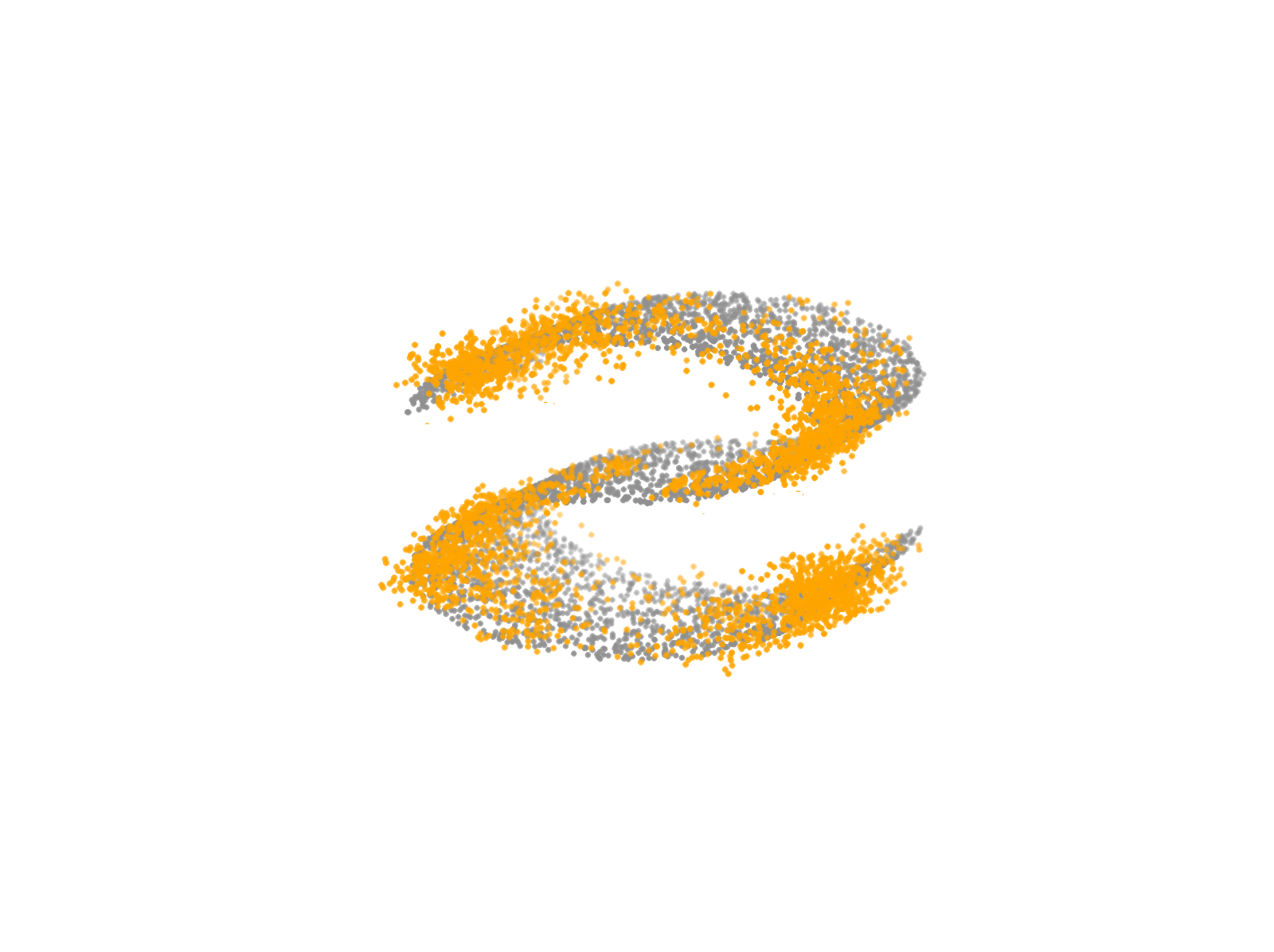}} \quad
\subfigure[Data space $\mathcal{X}$ for the Fish bowl dataset.]{\includegraphics[width=0.28\linewidth]{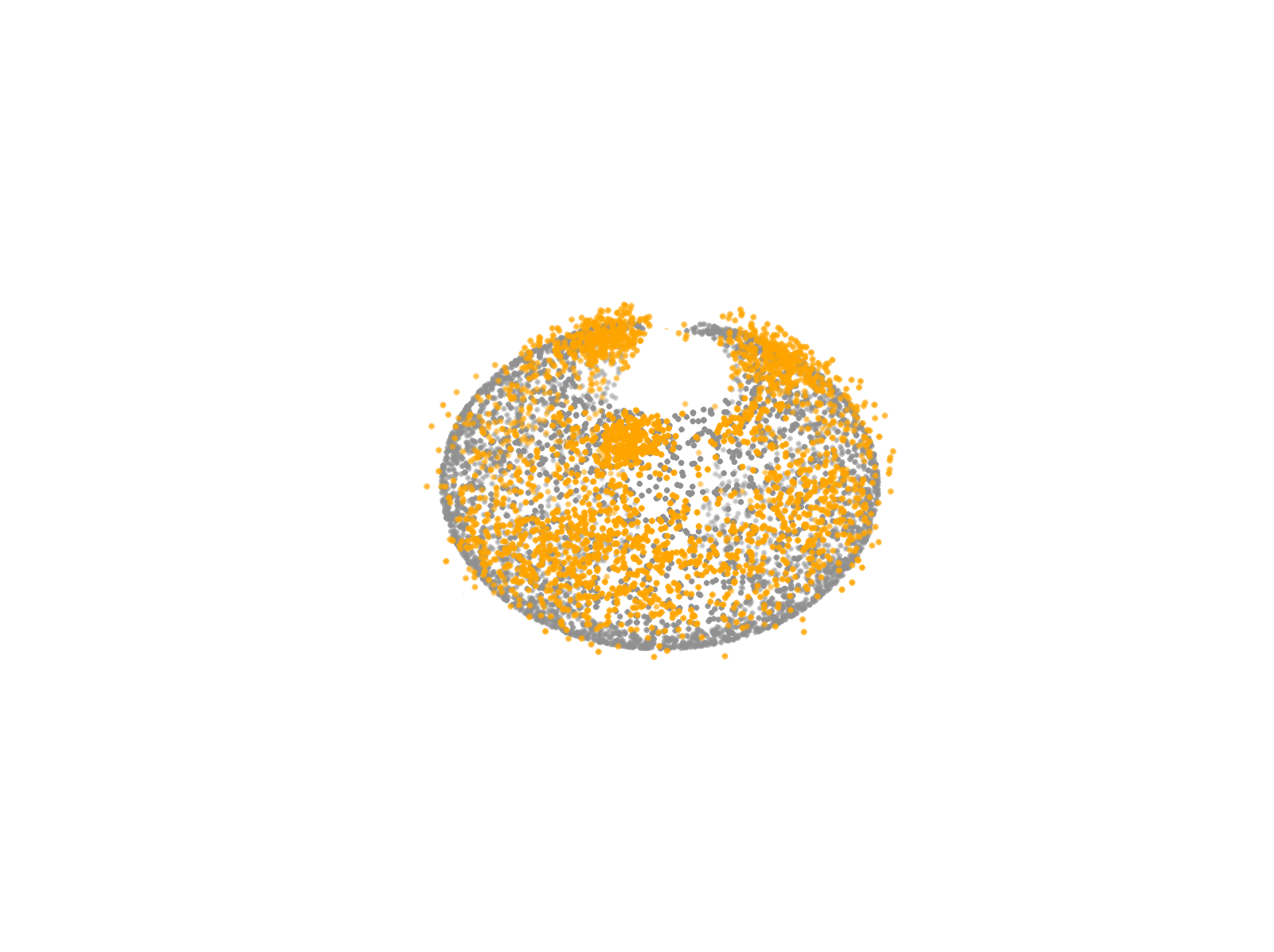}} 
\caption{Three toy $3$-D (gray data-points) datasets and corresponding samples drawn via CDE. CDE maps raw samples to latent features through a mapping ${\bm h}:\mathbb{R}^3 \rightarrow \mathbb{R}^2$ that is learned by an auto-encoder. The non-parametric latent density model allows efficient sample generation in the latent space (orange data-points on the images of the first row).  
The approximate inverse map of the decoder, back-transforms latent samples into samples in the data space (orange data-points on the images of the second row. See text for further details.}
\label{sec:2d_data}
\end{center}

\end{figure*}
We begin with modeling the joint density function of a subset of MNIST images, consisting of only $0$s and $8$s using the proposed CDE model. For these experiments the network architecture considered is a four-layer network encoder of $784$, $128$, $64$, $32$, neurons respectively (with the decoder being a mirrored version of the encoder), and ReLU activation functions. In Figure \ref{vae-comp}, we visualize random samples learned by the proposed model for different values of $F$ and $K$ to show that only a few parameters are needed to obtain a model that is flexible enough to fit the distribution in great detail. The first row represents results for fixed $F=4$ and different values of $K\in [1,3,5]$, while the second row represents results for fixed $K=3$ and different values of $F\in [2,4,8]$. Increasing $K$  generates sharper digits, while increasing $F$ better differentiates the samples of the two digits.

We then continue with modeling three toy $3$-D datasets, namely Swiss-roll, S-curve, and Fish-bowl where we are given $3000$ training data points from each dataset and we use CDE to randomly sample $5000$ synthetic data points. For all the datasets considered, the auto encoder structure we use is FC ($3$, $128$, ReLU), FC ($128$, $64$, ReLU), FC ($64$, $32$, ReLU), FC ($32$, 2, none), FC ($2$, $32$, ReLU), FC ($32$, $64$, ReLU), FC ($64$, $128$, ReLU), FC ($128$, $3$, none), and the tensor parameters are set to $(K,F)=(5,10)$. We provide an illustration of the $2$-D latent space learned via  latent synthetic samples drawn using the proposed method. Using the approximate inverse map of the decoder, we can back-transform latent samples into samples in the original data space and visualize the learned distribution in the original space. The results in Figure \ref{sec:2d_data} showcase that the proposed framework is capable of learning the structure of the data, notably in critical regions where the curvature is very high -- which is interesting. 

 \subsection{Tabular Datasets} 
 \begin{table*}[h!]
\begin{center}
\scalebox{1.0}{
\begin{tabular}{l|c|c|c|c|c|c|c|c|c|}
\toprule
\textbf{Data set}&\textbf{N}&\textbf{M}&\textbf{VAE} &\textbf{Real-NVP} &\textbf{MAF} &\textbf{GF}&\textbf{CDE} 
\\
\midrule
\textbf{MINIBOONE} &$51$&$130065$&$3.69\pm0.68$&$3.18\pm0.16$&$3.17\pm0.45$&$\mathbf{3.15\pm0.52}$&$\mathbf{3.12\pm0.43}$\\
\textbf{BSDS300} &$63$&$50000$&$0.37\pm0.08$&$0.60\pm0.02$ &$\mathbf{0.32\pm0.03}$&$0.48\pm0.03$&$\mathbf{0.30\pm0.02}$\\
\textbf{Gas Sensor}  &$128$&$13910$&$1.46\pm0.04$&$1.31\pm0.31$&$ 1.23\pm0.44$&$1.23\pm0.30$&$\mathbf{1.21\pm0.84}$\\
\textbf{Musk}  &$168$ &$6598$&$0.40\pm0.51$&$0.22\pm0.68$& $\mathbf{0.12\pm0.07}$& $0.19\pm0.08$&$\mathbf{0.13\pm0.23}$\\
\textbf{IDA2016Challenge} &$171$&$76000$&$0.23\pm0.06$&$0.18\pm0.09$&$\mathbf{0.12\pm0.05}$&$\mathbf{0.10\pm0.09}$&$\mathbf{0.11\pm0.16}$\\
\textbf{BlogFeedback}    &281&60021&$2.43\pm0.22$&$\mathbf{2.35\pm0.21}$&${2.37\pm0.22}$&$2.37\pm0.42$&$\mathbf{2.32\pm0.17}$\\
\textbf{ISOLET} &617&7797&$1.41\pm0.31$&$\mathbf{1.09\pm0.04}$&$1.11\pm0.38$&$1.85\pm0.21$&$\mathbf{1.03\pm0.56}$\\
\bottomrule
\end{tabular}
}
\end{center}
\caption{Dataset information and test-set MAE on UCI datasets. For each test sample, we choose the response variable $Y$ at random and estimate using Stochastic Gradient Ascent.}
\label{table:method_evaluation}
\end{table*}
\begin{table*}
\end{table*}
Next, we evaluate the proposed approach on several regression tasks using tabular datasets described in Table \ref{table:method_evaluation}. We compare our approach against standard baselines. For each dataset we split the data in two subsets: $80\%$ used for training and $20\%$ used for testing. The parameters for each method are chosen using $5$-fold cross-validation and we report the Mean Absolute Error (MAE) for the unseen data samples.
The results underline the superior performance of the proposed method for inference tasks. Regarding the auto-encoder's parameters for the datasets, the number of hidden layers varies according to the dataset dimensionality -- from three (MINIBOONE dataset) to six (ISOLET dataset). The most critical one is the hidden layer dimensionality $D \in \left\{16,32,48,64\right\}$, and concerning tensor parameters, the important ones are the tensor rank $F  \in \left\{5,10,20,30,50\right\}$ and the smoothing parameter $K  \in \left\{5,10,15,20,30\right\}$. The learning, drop-out rates, and regularization parameters were sampled from a uniform distribution in the range [$0.05,0.2$]. That is, we randomly sampled parameters from this range using a uniform distribution and used cross-validation to select the best set of parameters. The initial weight matrices were all sampled from the uniform distribution within the range [$-1,1$].  

We used Adam optimizer \cite{kingma2014adam} with a batch size of $500$. Overall, we observe that CDE outperforms the baselines on almost all datasets, and performs comparable to the winning method in the remaining ones. More specifically, CDE shows significantly lower test-set MAE on MINIBOONE, Gas Sensor, and BlogFeedback dataset compared to Real-NVP and MAF, which appear to be the next best performing models. Additionally CDE has a clear lead against the VAE especially on Gas Sensor, IDA2016Challenge, and ISOLET dataset, which confirms our initial motivation of using a non-parametric latent density estimator to improve model flexibility. 
\subsection{Image Datasets}
\begin{figure*}[h!]
    \centering
    \begin{subfigure}{}
        \includegraphics[width=0.3\linewidth]{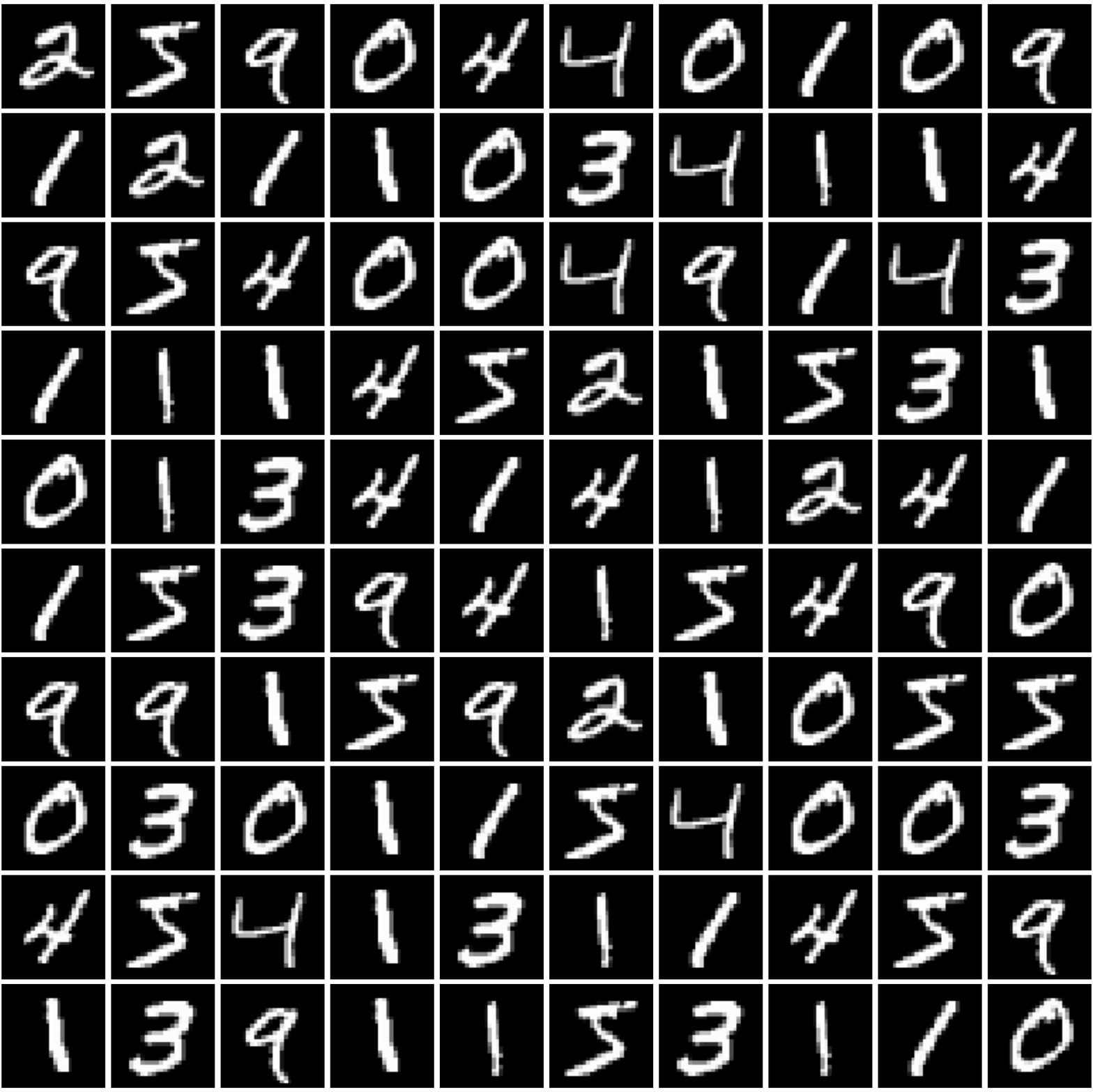}
    \end{subfigure}
            \begin{subfigure}{}
        \includegraphics[width=0.3\linewidth]{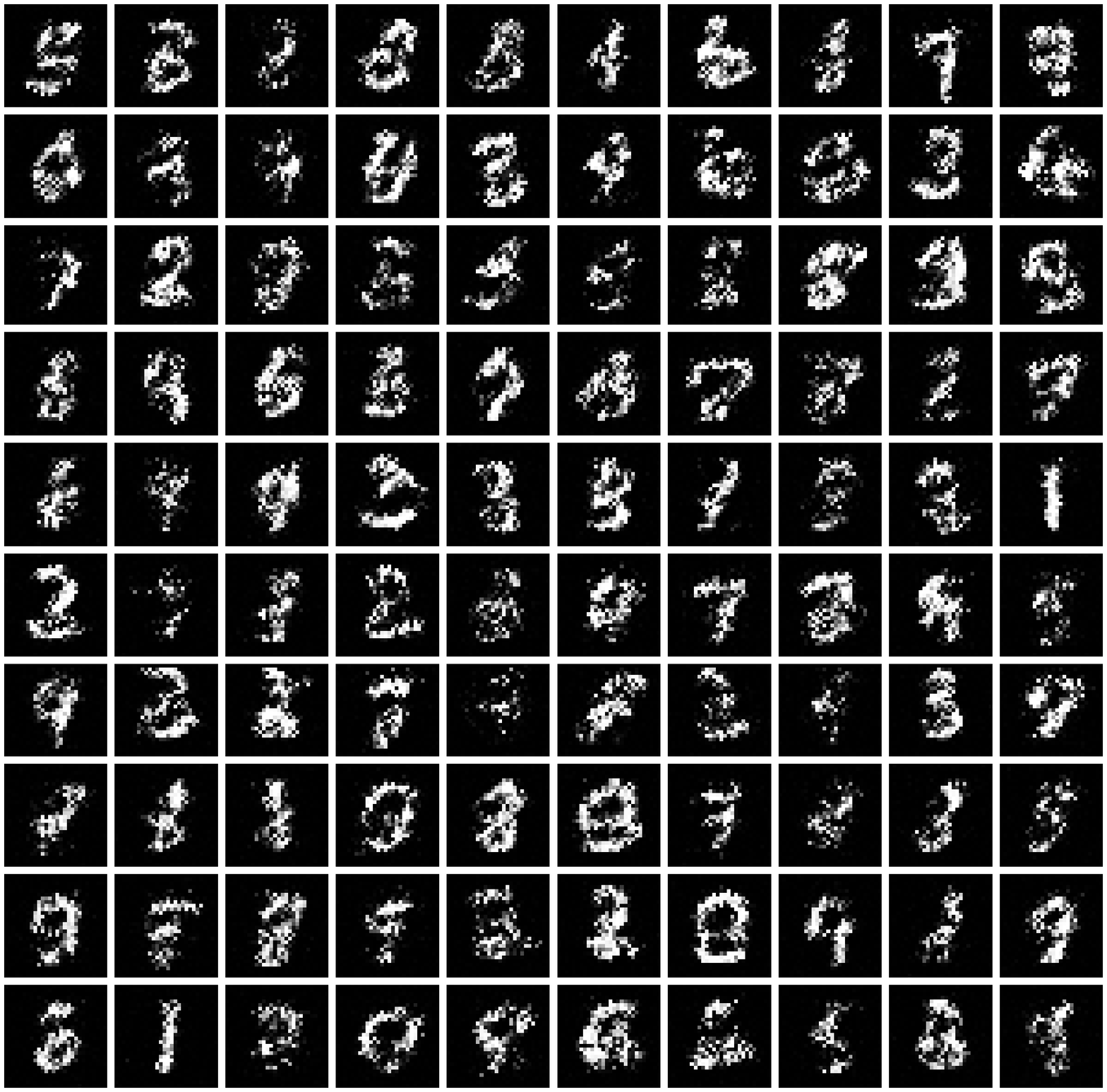}
    \end{subfigure}
    \begin{subfigure}{}
        \includegraphics[width=0.3\linewidth]{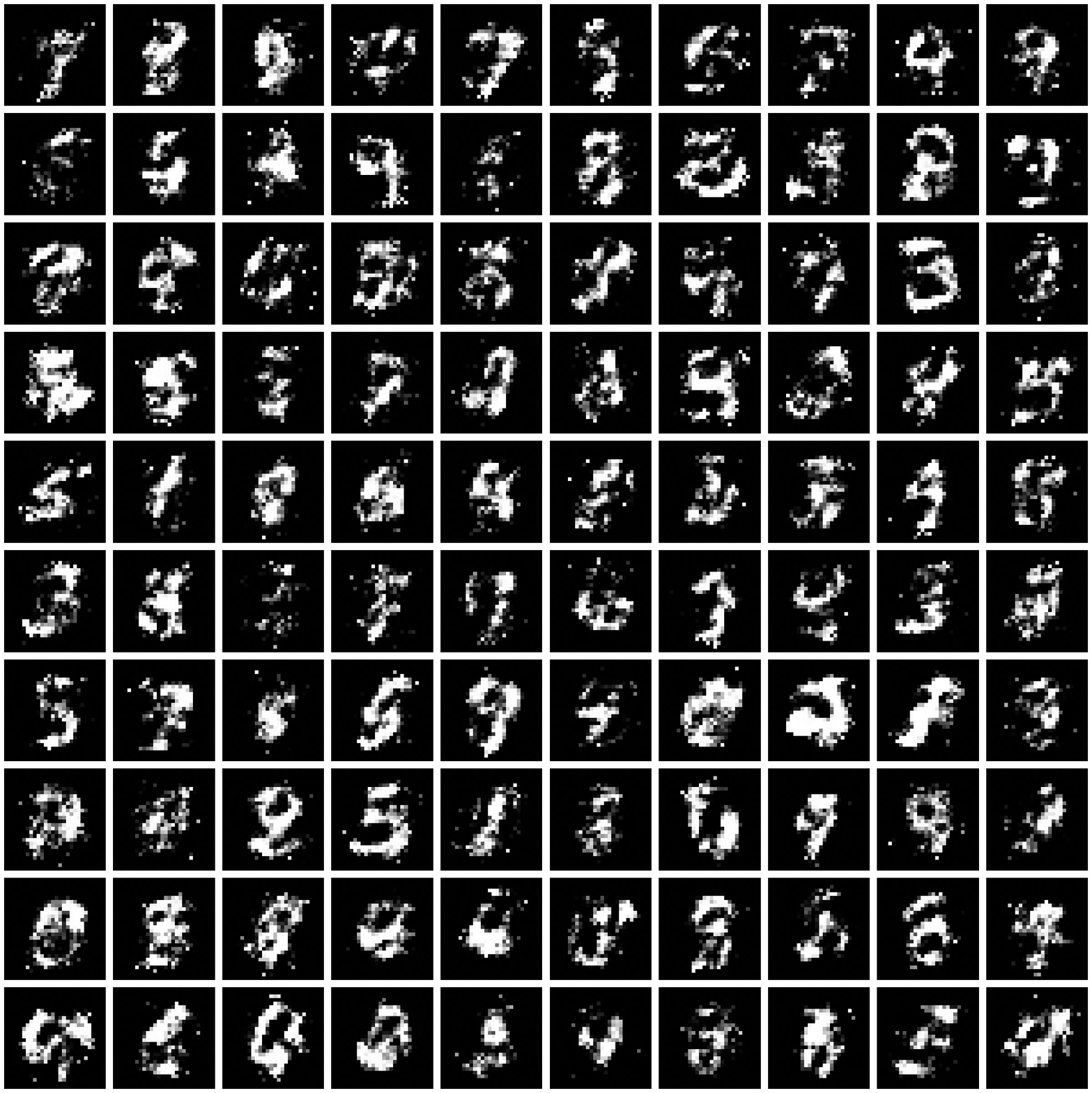}
    \end{subfigure}
    \\
        \begin{subfigure}{}
        \includegraphics[width=0.3\linewidth]{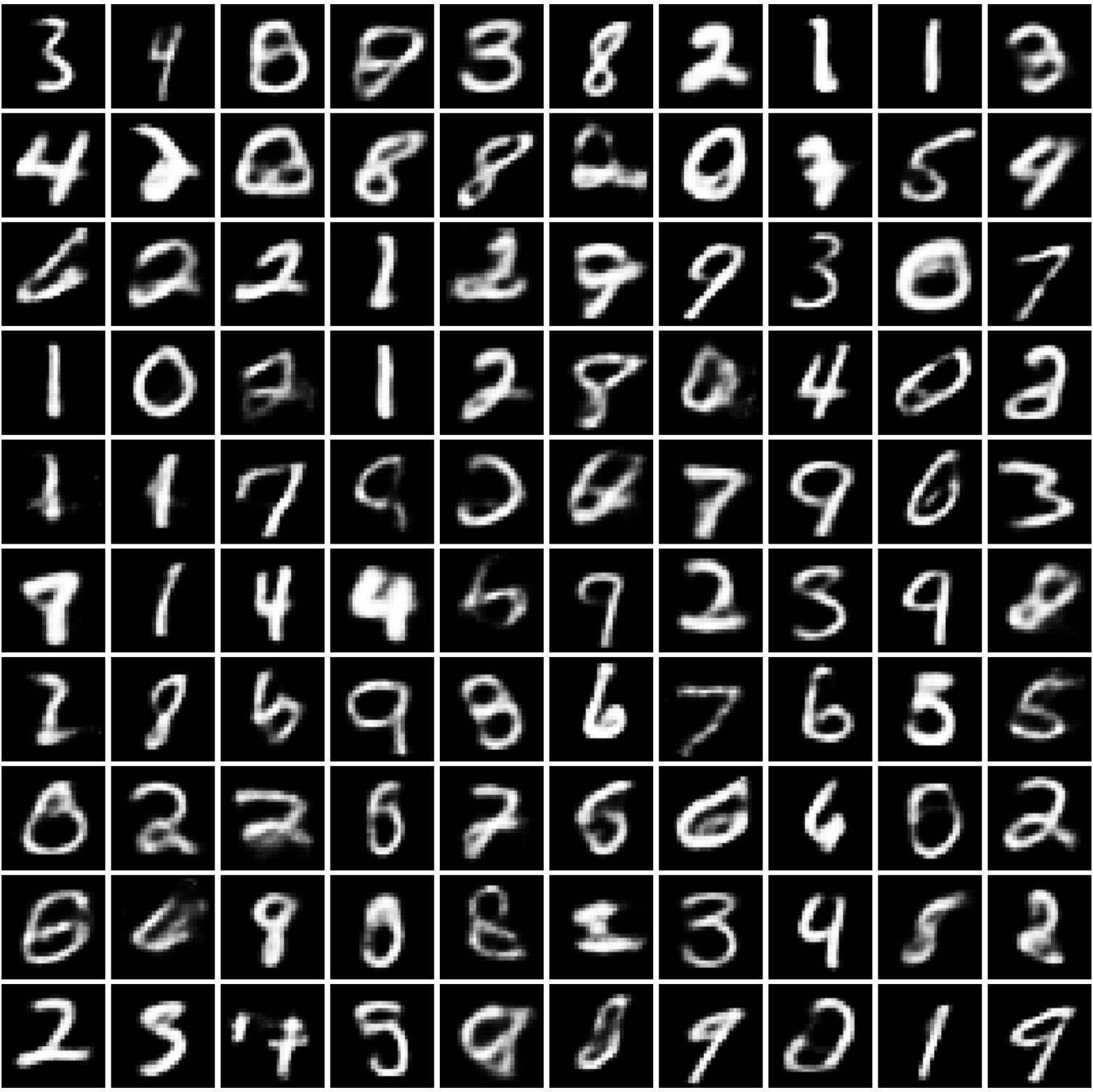}
    \end{subfigure}
        \begin{subfigure}{}
        \includegraphics[width=0.3\linewidth]{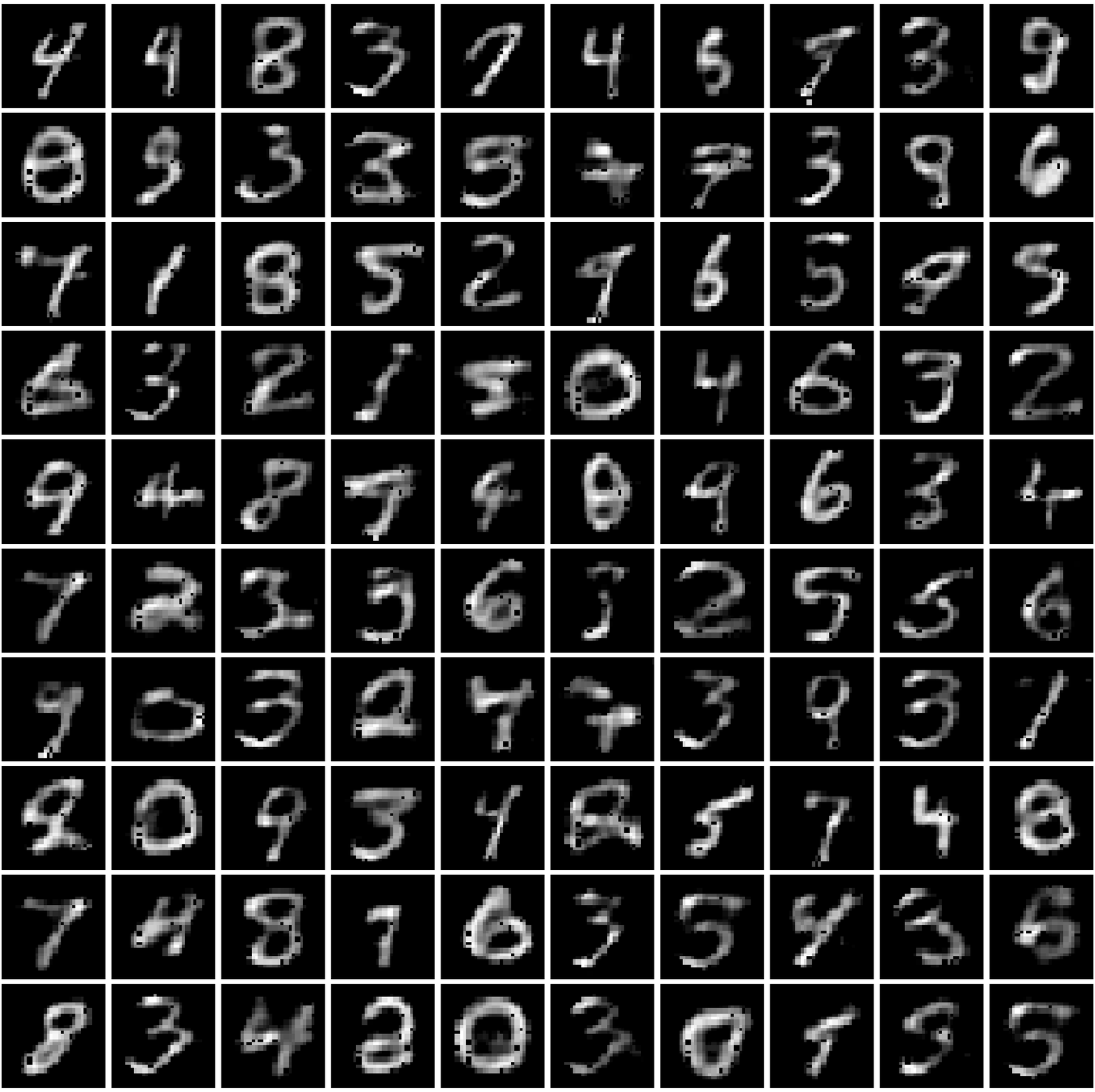}
    \end{subfigure}
\caption{Synthetic samples drawn from the joint density of MNIST from various models. From left to right: Ground Truth,  Masked auto-encoder for Distribution Estimation (MADE), Gaussianization Flows (GF), Variational auto-encoder (VAE), \textbf{Proposed: CDE}.} 
    \label{fig:mnist_samples}
\end{figure*}

\begin{figure*}[h!]

\bigskip
    \centering
    \begin{subfigure}{}
        \includegraphics[width=0.3\linewidth]{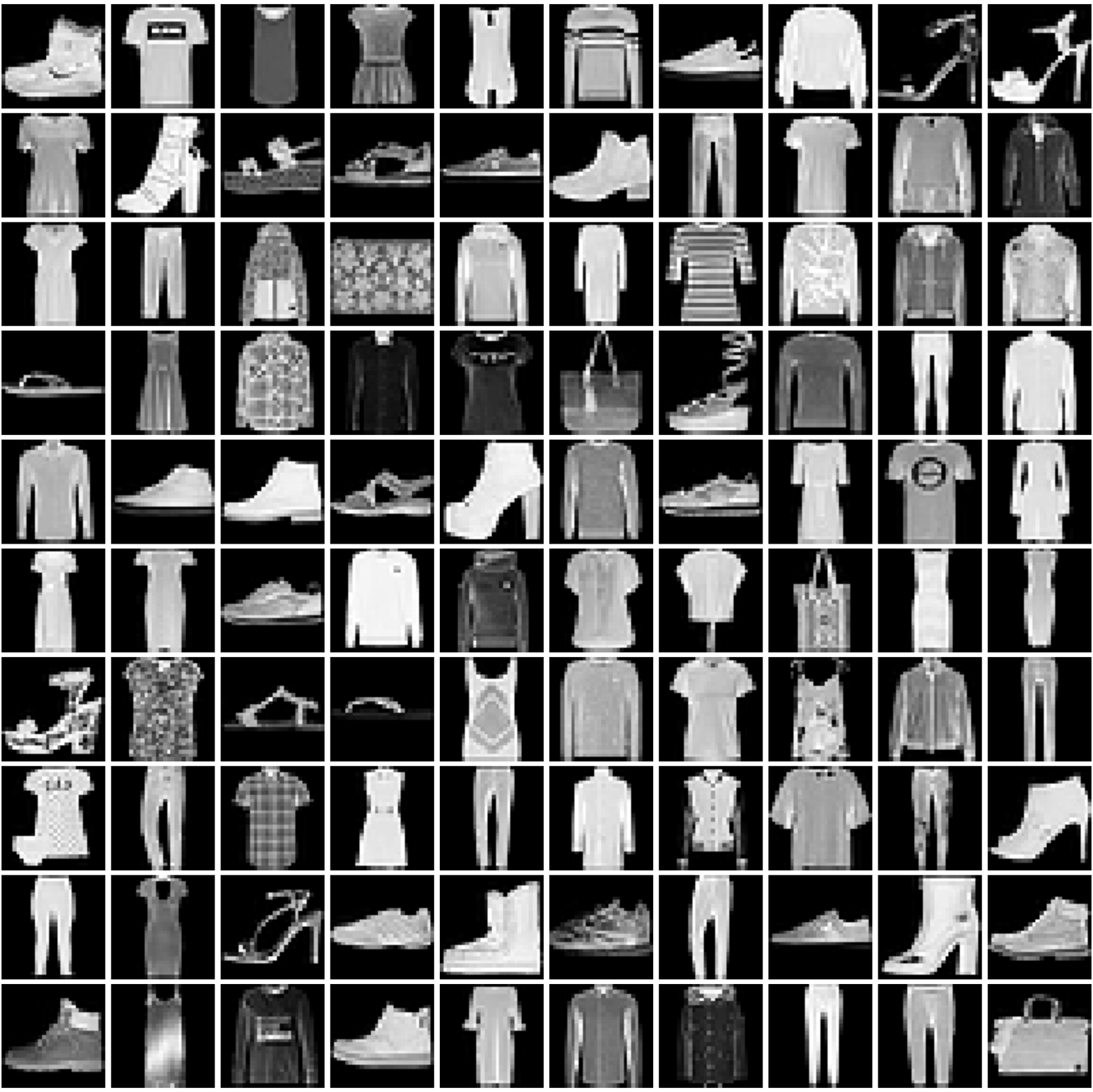}
    \end{subfigure}
        \begin{subfigure}{}
        \includegraphics[width=0.3\linewidth]{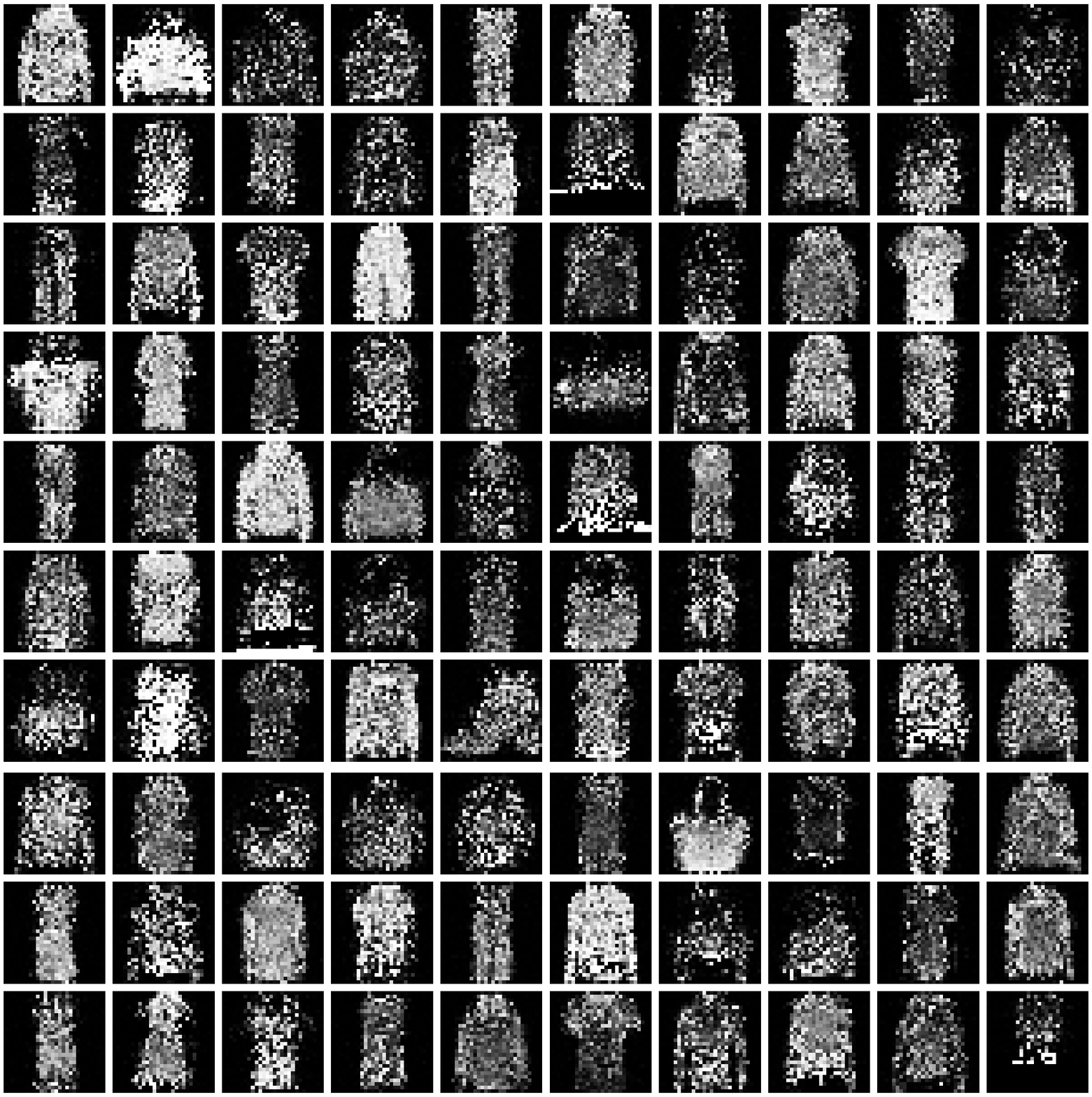}
    \end{subfigure}
    \begin{subfigure}{}
        \includegraphics[width=0.3\linewidth]{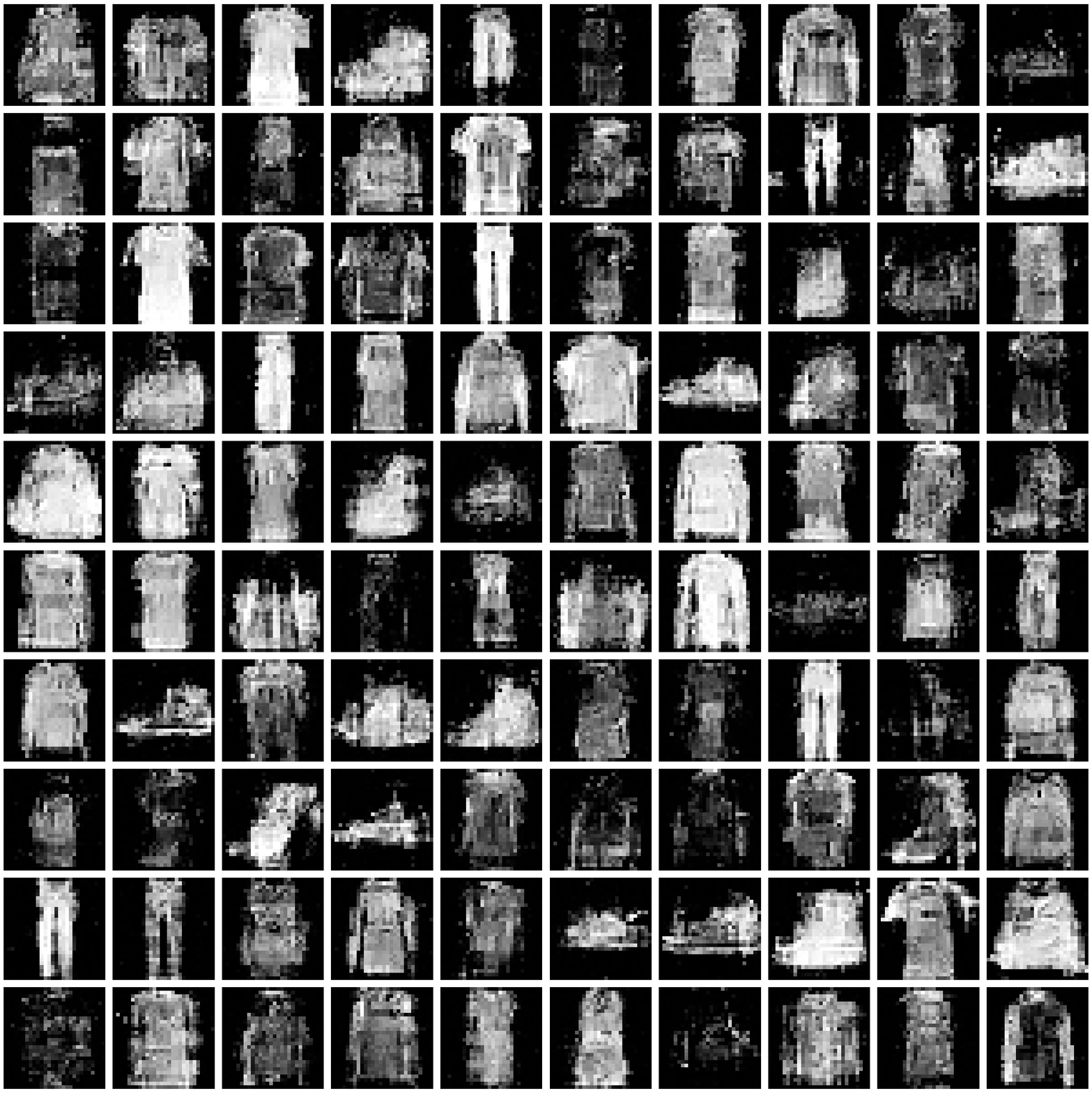}
    \end{subfigure}
    \\
        \begin{subfigure}{}
        \includegraphics[width=0.3\linewidth]{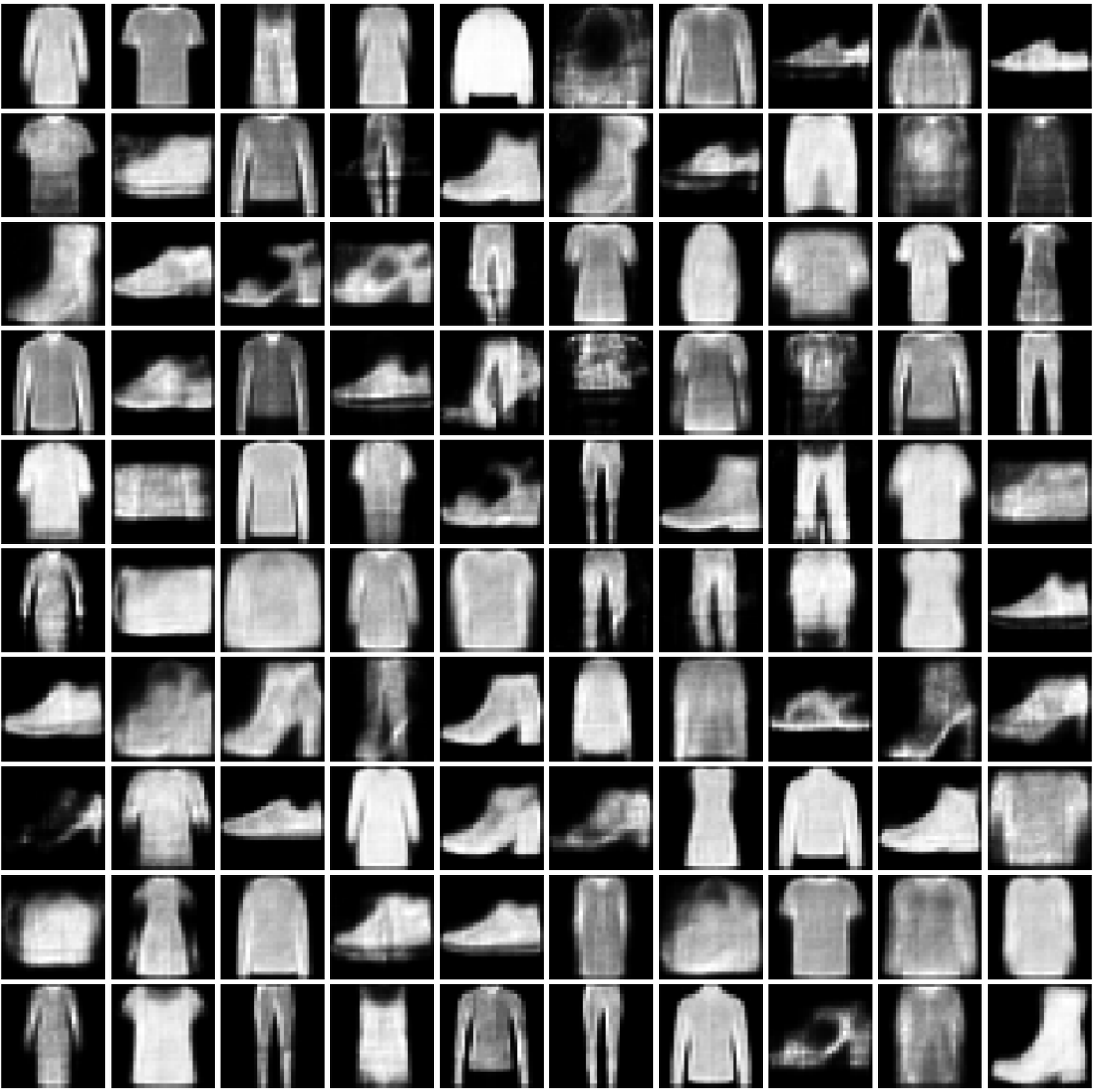}
    \end{subfigure}
    \begin{subfigure}{}
        \includegraphics[width=0.3\linewidth]{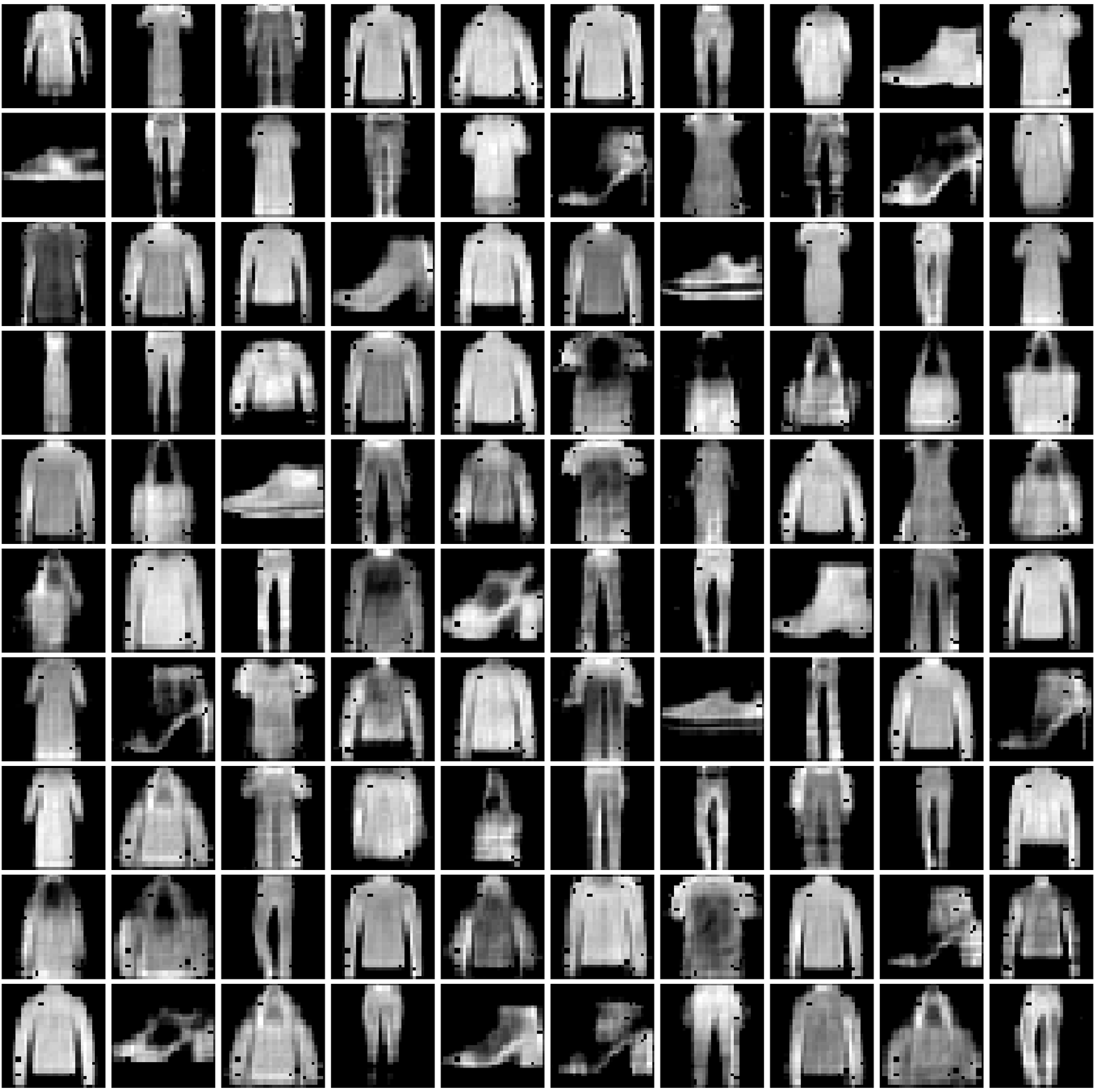}
    \end{subfigure}
\caption{Synthetic samples drawn from the joint density of Fashion-MNIST from various models. From left to right: Ground Truth,  Masked auto-encoder for Distribution Estimation (MADE), Gaussianization Flows (GF), Variational auto-encoder (VAE), \textbf{Proposed: CDE}.} 
    \label{fig:fmnist_samples}
\end{figure*}
We consider grayscale images from the MNIST and Fashion-MNIST database, which both contain a set of $60,000$ training observations of $28\times28$ pixels ($N=784$) from $10$ classes. Regarding MNIST, the most critical parameters include the encoder architecture, which consists of four hidden layers of  $784$, $128$, $64$, $32$, neurons respectively (the decoder network has a mirrored structure), ReLU activation function, tensor rank which is fixed to $F=50$, smoothing parameter $K=5$, and the learning rate which is fixed to $\alpha =0.0001$.
\begin{figure*}
    \begin{center}
        \includegraphics[width=0.35\linewidth]{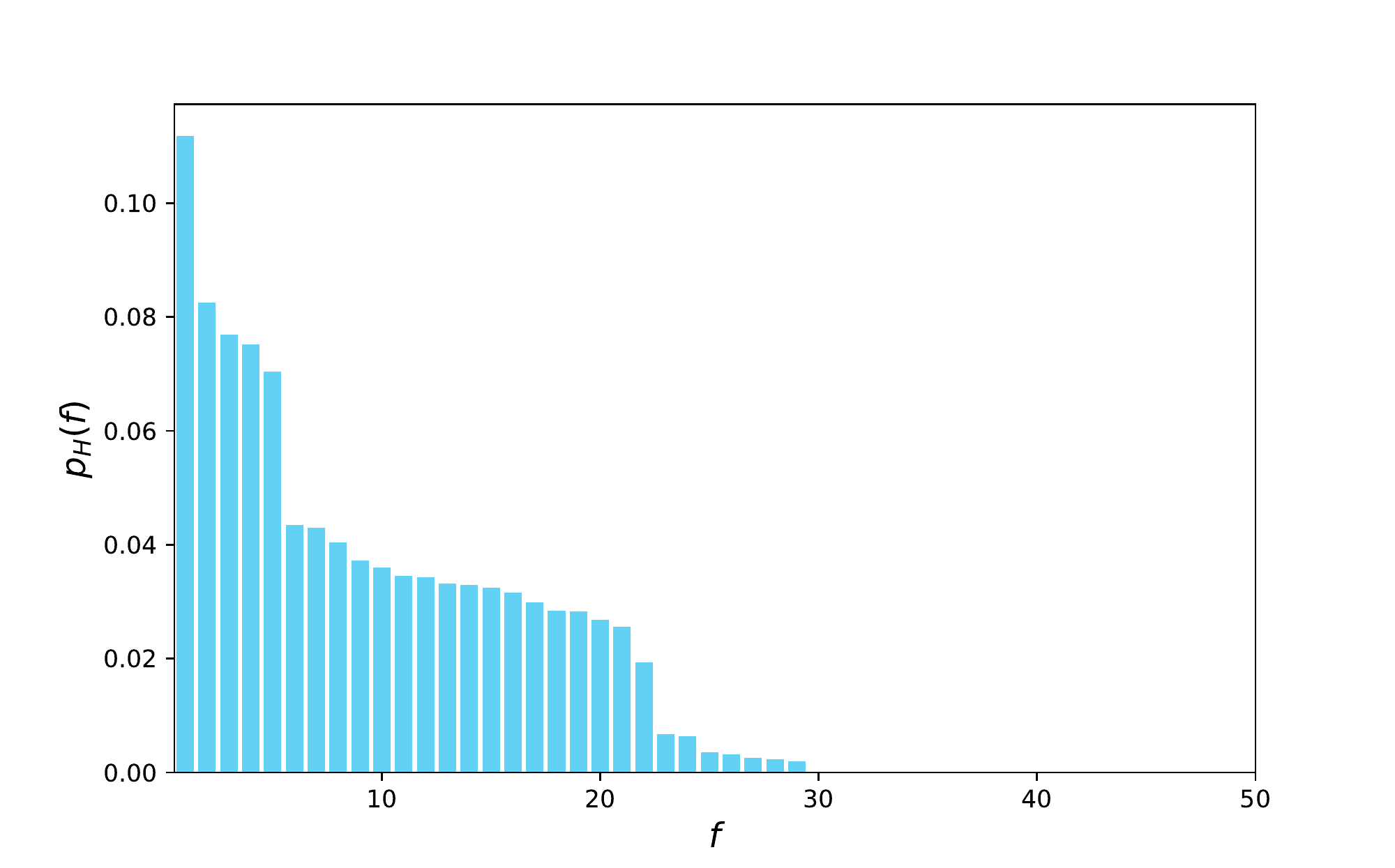}
        \includegraphics[width=0.35\linewidth]{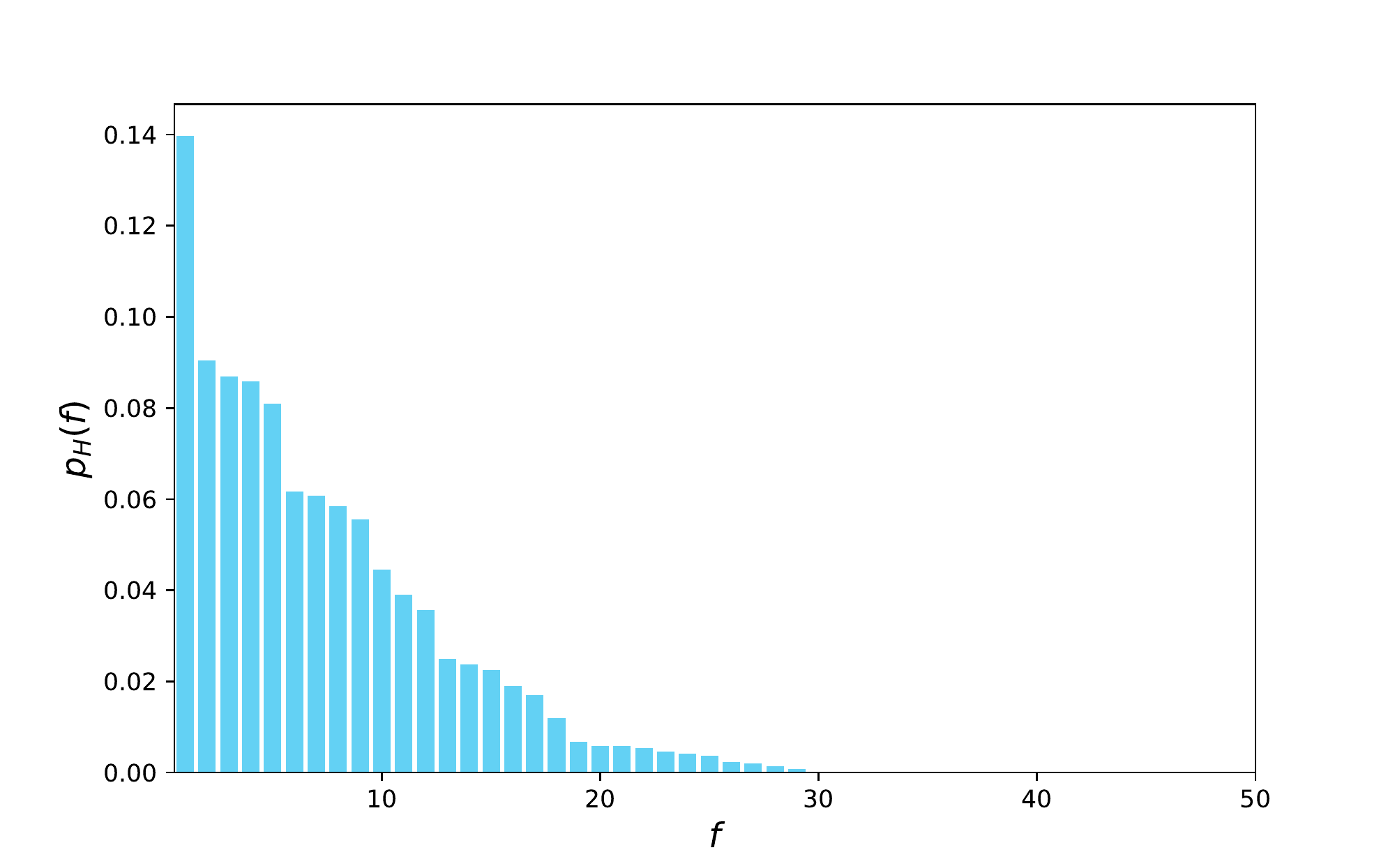}
         \end{center}
                 \caption{Distribution of the components of the latent variable $H$ after training the generative model on MNIST (left figure) and Fashion-MNIST (right figure).}
         \label{mnist_lambda}
    \end{figure*}
For Fashion-MNIST, the model parameters are fixed to be the same as for the MNIST dataset, with the encoder network $(784, 256, 128, 64, 32, 16)$ being the only exception. See also Figure \ref{mnist_lambda}, which shows the distribution of the components of the latent variable $H$ after training the generative model. These bar-plots tell us that the rank of the compressed density model is essentially $F=30$, but for exploratory modeling purposes, we set rank the to $F=50$ and encourage sparsity of the latent components through our optimization problem formulation. We sample from the learned lower-dimensional latent joint generative model ($D=32$ for MNIST and $D=16$ for Fashion-MNIST) -- See Section \ref{sec:Main_method} for the detailed sampling process -- and provide visualization of the generated data. The resulting $100$ randomly drawn samples, which are impressively more pleasing to the eye in direct comparison with other well-known models such as MADE and GFs are shown in Figures \ref{fig:mnist_samples} and \ref{fig:fmnist_samples}.

Although at first glance, the images generated by VAE have cleaner and thicker strokes, they are also more blurry and distorted than those produced by CDE. The Fashion-MNIST data help bring this out more clearly: one can see that CDE allows capturing and representing more details in the items, while the samples drawn from the VAE are much more blurry. The overall conclusion from the experiments is that while the quality of our synthetic images is competitive against the VAE and considerably better than that of the samples generated by the rest of the models considered, the proposed framework is superior for regression tasks.  
\subsection{Anomaly Detection Using Real Data}\label{exp_ano}
\renewcommand{\thefootnote}{\fnsymbol{footnote}}
\renewcommand{\thempfootnote}{\fnsymbol{mpfootnote}}

We use four public datasets\footnote[3]{Datasets can be downloaded at \url{https://kdd.ics.uci.edu/} and \url{http://odds.cs.stonybrook.edu}.}: KDDCUP99, Thyroid, Arrhythmia, and KDDCUP-Rev. The (instance number $M$, dimension $N$, anomaly ratio (\%)) of each dataset is (494021, 121, 20),  (3772, 6, 2.5),  (452, 274, 15), and (121597, 121, 20).  For categorical features, we further used one-hot representation to encode them. Regarding Thyroid, there are three classes in the original dataset. We treat the hyperfunction class as the anomaly class and the other two classes are treated as normal class. Regarding Arrhythmia, the smallest classes, including $3, 4, 5, 7, 8, 9, 14,$ and $15,$ are combined to form
the anomaly class, and the rest of the classes are combined to form the normal class. We randomly extracted 50\% of the data and assigned it to the training subset and the rest to the testing subset. In our experimental setting for anomaly detection, clean training data is adopted  -- that is, during the training, only normal data were used. We assume that the percentage of anomalous data points is known, and our goal is to detect which data points in the testing subset are most likely to be outliers. Towards this end, at the testing stage the likelihood of each testing sample in the compressed domain was evaluated and sorted in order to detect the anomalies as the points with the smallest likelihood. By knowing the percentage of anomalies, we can indicate the exact number of outliers and output the data samples with the smallest likelihood values. 

The network structures of CDE used for  individual datasets are summarized as follows.
\begin{itemize}
\item For KDDCUP, the auto-encoder network runs with FC (120, 60, tanh), FC (60, 30, tanh), FC (30, 20, tanh), FC (20, 10, none), FC (10, 20, tanh), FC (20, 30, tanh), FC (30, 60, tanh), FC (60, 120, none).

\item The auto-encoder network for Thyroid runs with FC (6, 12, tanh), FC (12, 4, tanh), FC (4, 2, none), FC (2, 4, tanh), FC (4, 12, tanh), FC (12, 6, none).

\item The auto-encoder network for Arrhythmia runs with FC (274, 64, tanh), FC (64, 32, none), FC (32, 64, tanh), FC (64, 274, none).

\item The auto-encoder network for KDDCUP-Rev runs with FC (120, 60, tanh), FC (60, 30, tanh), FC (30, 20, tanh), FC (20, 10, none), FC (10, 20, tanh), FC (20, 30, tanh), FC (30, 60, tanh), FC (60, 120, none).

\begin{table*}[t]
\begin{center}
\renewcommand{\footnoterule}{\empty}
\begin{minipage}{\textwidth}
\begin{center}
\begin{tabular}{c|l|lll}
\multicolumn{1}{l|}{Dataset} & Methods       & Precision      & Recall         & F1             \\ \hline
\multirow{3}{*}{KDDCup}  & VAE         & 0.9524 (0.0047)         & 0.9140 (0.0052)        & 0.9326 (0.0052)        \\
                             & DAGMM         & 0.9427 (0.0055) & 0.9578 (0.0051) & 0.9507 (0.0052) \\
                             & CDE   & \bf{0.9565 (0.0046)} & \bf{0.9712 (0.0048)} & \bf{0.9641 (0.0045)} \\ \hline
\multirow{3}{*}{Thyroid}     & VAE         &  \bf{0.6575  (0.0371)}         & 0.5743  (0.0583)        & 0.6357  (0.0583)        \\
                             & DAGMM        & 0.4658 (0.0481) & 0.4902 (0.0452) & 0.4752 (0.0497) \\
                             & CDE   &  0.6560 (0.0572) &  \bf{0.6740 (0.0493)} &  \bf{0.6703 (0.0592)} \\ \hline
\multirow{3}{*}{Arrythmia}   & VAE        & 0.4375 (0.0538)        & 0.4340 (0.0496)        & 0.4302 (0.0482)        \\
                             & DAGMM    &  \bf{0.5358 (0.0468)} &  \bf{0.5592 (0.0475)} &  \bf{0.5403 (0.0421)} \\
                             & CDE   & 0.5299 (0.0400) & 0.5551 (0.0418) & 0.5389 (0.0420) \\ \hline
\multirow{3}{*}{KDDCup-rev} 
                             & VAE       & 0.9771 (0.0058) & 0.9779 (0.0004) & 0.9678 (0.0018)\\
                             & DAGMM    & 0.9762 (0.0038) & 0.9823 (0.0017) & 0.9709 (0.0021) \\
                             & CDE   &  \bf{0.9866 (0.0008)} &  \bf{0.9872 (0.0015)} &  \bf{0.9871 (0.0012)}\\ \hline
\end{tabular}
\end{center}
\end{minipage}
\end{center}
\caption{Average (over 20 runs) and standard deviations (in brackets) of Precision, Recall and F1 score}\label{tab:anomaly} 
\end{table*}
\end{itemize} 
As metrics, precision, recall, and F1 score are calculated. We run experiments 20 times for each dataset split by 20 different random seeds. Table \ref{tab:anomaly} reports the average scores and standard deviations (in brackets). Compared to the baselines considered, CDE achieves the highest performance -- CDE is superior to both VAE and DAGMM 
for each evaluation criterion except for precision on the Arrhythmia dataset. This result suggests that our proposed latent nonparametric density estimation approach can provide more expressive models which can bring better performance in important detection tasks as well. 

\section{Conclusions}
In this work, we introduced Compressed Density Estimation (CDE), a novel probabilistic latent density model that builds upon deep auto-encoder networks and non-parametric multivariate density modeling in the Fourier domain. We propose using an auto-encoder to embed the data into a latent code space by minimizing reconstruction error, and a regularization over the latent space which maximizes the likelihood of the hidden code vector and is modelled using a low-rank characteristic tensor approach.

We investigated whether leveraging probabilistic (non-parametric) low-rank tensor models in the Fourier domain  as a latent distribution model can improve the expressivity of density models. By jointly optimizing the auto-encoder and the latent density model, we can better capture the latent distribution of data representations obtained by the auto-encoder. Experimental results demonstrated the effectiveness of the proposed joint optimization approach, which is able to learn complex high dimensional distributions using a parsimonious model with few tuning parameters.

\bibliographystyle{IEEEtran}
\bibliography{references}

\begin{thebibliography}{10}
\providecommand{\url}[1]{#1}
\csname url@samestyle\endcsname
\providecommand{\newblock}{\relax}
\providecommand{\bibinfo}[2]{#2}
\providecommand{\BIBentrySTDinterwordspacing}{\spaceskip=0pt\relax}
\providecommand{\BIBentryALTinterwordstretchfactor}{4}
\providecommand{\BIBentryALTinterwordspacing}{\spaceskip=\fontdimen2\font plus
\BIBentryALTinterwordstretchfactor\fontdimen3\font minus
  \fontdimen4\font\relax}
\providecommand{\BIBforeignlanguage}[2]{{%
\expandafter\ifx\csname l@#1\endcsname\relax
\typeout{** WARNING: IEEEtran.bst: No hyphenation pattern has been}%
\typeout{** loaded for the language `#1'. Using the pattern for}%
\typeout{** the default language instead.}%
\else
\language=\csname l@#1\endcsname
\fi
#2}}
\providecommand{\BIBdecl}{\relax}
\BIBdecl

\bibitem{kingma2018glow}
D.~P. Kingma and P.~Dhariwal, ``Glow: Generative flow with invertible 1x1
  convolutions,'' in \emph{Advances in Neural Information Processing Systems},
  2018, pp. 10\,215--10\,224.

\bibitem{oord2016wavenet}
A.~v.~d. Oord, S.~Dieleman, H.~Zen, K.~Simonyan, O.~Vinyals, A.~Graves,
  N.~Kalchbrenner, A.~Senior, and K.~Kavukcuoglu, ``Wavenet: A generative model
  for raw audio,'' \emph{arXiv preprint arXiv:1609.03499}, 2016.

\bibitem{bowman2015generating}
S.~R. Bowman, L.~Vilnis, O.~Vinyals, A.~M. Dai, R.~Jozefowicz, and S.~Bengio,
  ``Generating sentences from a continuous space,'' \emph{arXiv preprint
  arXiv:1511.06349}, 2015.

\bibitem{zong2018deep}
B.~Zong, Q.~Song, M.~R. Min, W.~Cheng, C.~Lumezanu, D.~Cho, and H.~Chen, ``Deep
  autoencoding gaussian mixture model for unsupervised anomaly detection,'' in
  \emph{International Conference on Learning Representations}, 2018.

\bibitem{silverman1986density}
B.~W. Silverman, \emph{Density estimation for statistics and data
  analysis}.\hskip 1em plus 0.5em minus 0.4em\relax CRC press, 1986, vol.~26.

\bibitem{mclachlan2004finite}
G.~J. McLachlan and D.~Peel, \emph{Finite mixture models}.\hskip 1em plus 0.5em
  minus 0.4em\relax John Wiley \& Sons, 2004.

\bibitem{goodfellow2014generative}
I.~Goodfellow, J.~Pouget-Abadie, M.~Mirza, B.~Xu, D.~Warde-Farley, S.~Ozair,
  A.~Courville, and Y.~Bengio, ``Generative adversarial nets,'' in
  \emph{Advances in Neural Information Processing Systems}, 2014, pp.
  2672--2680.

\bibitem{kingma2013auto}
D.~P. Kingma and M.~Welling, ``Auto-encoding variational bayes,'' in
  \emph{International Conference on Learning Representations}, 2014.

\bibitem{dai2019diagnosing}
B.~Dai and D.~Wipf, ``Diagnosing and enhancing vae models,'' \emph{arXiv
  preprint arXiv:1903.05789}, 2019.

\bibitem{rosca2018distribution}
M.~Rosca, B.~Lakshminarayanan, and S.~Mohamed, ``Distribution matching in
  variational inference,'' \emph{arXiv preprint arXiv:1802.06847}, 2018.

\bibitem{oord2016pixel}
A.~V. Oord, N.~Kalchbrenner, and K.~Kavukcuoglu, ``Pixel recurrent neural
  networks,'' in \emph{International Conference on Machine Learning}, vol.~48,
  20--22 Jun 2016, pp. 1747--1756.

\bibitem{dinh2014nice}
L.~Dinh, D.~Krueger, and Y.~Bengio, ``{NICE}: Non-linear independent components
  estimation,'' \emph{arXiv preprint arXiv:1410.8516}, 2015.

\bibitem{dinh2016density}
L.~Dinh, J.~Sohl-Dickstein, and S.~Bengio, ``Density estimation using real
  {NVP},'' in \emph{International Conference on Learning Representations},
  2016.

\bibitem{ho2019flow++}
J.~Ho, X.~Chen, A.~Srinivas, Y.~Duan, and P.~Abbeel, ``Flow++: Improving
  flow-based generative models with variational dequantization and architecture
  design,'' in \emph{International Conference on Machine Learning}, 2019, pp.
  2722--2730.

\bibitem{amiridi2020nonparametric}
M.~Amiridi, N.~Kargas, and N.~D. Sidiropoulos, ``Nonparametric multivariate
  density estimation: A low-rank characteristic function approach,''
  \emph{arXiv preprint arXiv:2008.12315}, 2020.

\bibitem{Harshman1970}
R.~A. Harshman, ``Foundations of the {PARAFAC} procedure: Models and conditions
  for an ``explanatory" multimodal factor analysis,'' \emph{UCLA Working Papers
  Phonetics}, vol.~16, pp. 1--84, 1970.

\bibitem{lecun1998mnist}
Y.~LeCun, ``The {MNIST} database of handwritten digits,'' \emph{http://yann.
  lecun. com/exdb/mnist/}, 1998.

\bibitem{xiao2017fashion}
H.~Xiao, K.~Rasul, and R.~Vollgraf, ``Fashion-mnist: a novel image dataset for
  benchmarking machine learning algorithms,'' \emph{arXiv preprint
  arXiv:1708.07747}, 2017.

\bibitem{scott1991feasibility}
D.~W. Scott, ``Feasibility of multivariate density estimates,''
  \emph{Biometrika}, vol.~78, no.~1, pp. 197--205, 1991.

\bibitem{hoffman2016elbo}
M.~D. Hoffman and M.~J. Johnson, ``Elbo surgery: yet another way to carve up
  the variational evidence lower bound,'' in \emph{Workshop in Advances in
  Approximate Bayesian Inference, NIPS}, vol.~1, 2016, p.~2.

\bibitem{uria2013rnade}
B.~Uria, I.~Murray, and H.~Larochelle, ``{RNADE}: The real-valued neural
  autoregressive density-estimator,'' in \emph{Advances in Neural Information
  Processing Systems}, 2013, pp. 2175--2183.

\bibitem{germain2015made}
M.~Germain, K.~Gregor, I.~Murray, and H.~Larochelle, ``{MADE}: Masked
  autoencoder for distribution estimation,'' in \emph{International Conference
  on Machine Learning}, 2015, pp. 881--889.

\bibitem{rezende15}
D.~Rezende and S.~Mohamed, ``Variational inference with normalizing flows,'' in
  \emph{International Conference on Machine Learning}, vol.~37, 2015, pp.
  1530--1538.

\bibitem{papamakarios2017masked}
G.~Papamakarios, T.~Pavlakou, and I.~Murray, ``Masked autoregressive flow for
  density estimation,'' in \emph{Advances in Neural Information Processing
  Systems}, 2017, pp. 2338--2347.

\bibitem{meng2020gaussianization}
C.~Meng, Y.~Song, J.~Song, and S.~Ermon, ``Gaussianization flows,'' in
  \emph{International Conference on Artificial Intelligence and Statistics},
  2020, pp. 4336--4345.

\bibitem{hitchcock1927expression}
F.~L. Hitchcock, ``The expression of a tensor or a polyadic as a sum of
  products,'' \emph{Journal of Mathematics and Physics}, vol.~6, no. 1-4, pp.
  164--189, 1927.

\bibitem{sidiropoulos2017tensor}
N.~D. Sidiropoulos, L.~De~Lathauwer, X.~Fu, K.~Huang, E.~E. Papalexakis, and
  C.~Faloutsos, ``Tensor decomposition for signal processing and machine
  learning,'' \emph{IEEE Transactions on Signal Processing}, vol.~65, no.~13,
  pp. 3551--3582, 2017.

\bibitem{nair2010rectified}
V.~Nair and G.~E. Hinton, ``Rectified linear units improve restricted boltzmann
  machines,'' in \emph{ICML}, 2010.

\bibitem{lecun1998gradient}
Y.~LeCun, L.~Bottou, Y.~Bengio, and P.~Haffner, ``Gradient-based learning
  applied to document recognition,'' \emph{Proceedings of the IEEE}, vol.~86,
  no.~11, pp. 2278--2324, 1998.

\bibitem{krizhevsky2017imagenet}
A.~Krizhevsky, I.~Sutskever, and G.~E. Hinton, ``Imagenet classification with
  deep convolutional neural networks,'' \emph{Communications of the ACM},
  vol.~60, no.~6, pp. 84--90, 2017.

\bibitem{plonka2018numerical}
G.~Plonka, D.~Potts, G.~Steidl, and M.~Tasche, \emph{Numerical Fourier
  Analysis}.\hskip 1em plus 0.5em minus 0.4em\relax Springer, 2018.

\bibitem{mason1980near}
J.~C. Mason, ``Near-best multivariate approximation by fourier series,
  chebyshev series and chebyshev interpolation,'' \emph{Journal of
  Approximation Theory}, vol.~28, no.~4, pp. 349--358, 1980.

\bibitem{handscomb2014methods}
D.~C. Handscomb, \emph{Methods of numerical approximation: lectures delivered
  at a Summer School held at Oxford University, September 1965}.\hskip 1em plus
  0.5em minus 0.4em\relax Elsevier, 2014.

\bibitem{kingma2014adam}
D.~P. Kingma and J.~Ba, ``Adam: A method for stochastic optimization,''
  \emph{arXiv preprint arXiv:1412.6980}, 2014.

\end{thebibliography}

\end{document}